\documentclass[10pt]{article}

\usepackage{cite}

\usepackage{color}

\usepackage{graphicx}
\usepackage{amssymb,amsmath}
\usepackage{url}

\usepackage[labelfont=bf,labelsep=period,justification=raggedright]{caption}

\makeatletter
\renewcommand{\@biblabel}[1]{\quad#1.}
\makeatother

\date{}

\usepackage[utf8]{inputenc}
\begin{document}

\begin{flushleft}
{\Large
\textbf{Multi-objective analysis of computational models}
}
\\
Stephane Doncieux$^{1,\ast}$, 
Jean Liénard$^{1}$, 
Benoît Girard$^{1}$,
Mohamed Hamdaoui$^{2}$,
Jo\"el Chaskalovic$^{3}$,
\\
\bf{1} Institut des Systèmes Intelligents et de Robotique, Université Pierre et Marie Curie-Paris 6, CNRS UMR 7222, Paris, France
\\
\bf{2} Department of Applied Mathematics and Systems, Ecole Centrale Paris, Paris, France
\\
\bf{3} Institut Jean le Rond d'Alembert, Université Pierre et Marie Curie-Paris 6, CNRS UMR 7190, Paris, France
\\
$\ast$ E-mail: stephane.doncieux@isir.upmc.fr
\end{flushleft}

\section*{abstract}
Computational models are of increasing complexity and their behavior may in particular emerge from the interaction of different parts. Studying such models becomes then more and more difficult and there is a need for methods and tools supporting this process. Multi-objective evolutionary algorithms generate a set of trade-off solutions instead of a single optimal solution. The availability of a set of solutions that have the specificity to be optimal relative to carefully chosen objectives allows to perform data mining in order to better understand model features and regularities. We review the corresponding work, propose a unifying framework, and highlight its potential use. Typical questions that such a methodology allows to address are the following: what are the most critical parameters of the model? What are the relations between the parameters and the objectives? What are the typical behaviors of the model?  Two examples are provided to illustrate the capabilities of the methodology. The features of a flapping-wing robot are thus evaluated to find out its speed-energy relation, together with the criticality of  its parameters. A neurocomputational model of the Basal Ganglia brain nuclei is then considered and its most salient features according to this methodology are presented and discussed.



\section*{Nomenclature}
\begin{tabbing}
 XXXXX \= \kill
$\mathbf{X}$ \> parameter to be optimized ($\mathbf{X}\in \mathbb{R}^m$)\\
$f(\mathbf{X})$ \> function to be optimized\\
$\prec$ \> dominance relation\\
$\preceq$ \> dominance or equality relation\\
$\chi$ \> set of Pareto set approximations\\
$\Gamma$ \> set of independent runs of the optimizer\\
$\alpha_\chi(z)$ \> attainment function\\
$\Psi_1$ \> attainment surface corresponding to a value of 1 of the attainment function\\
$\Psi_0$ \> attainment surface corresponding to a value of 0 of the attainment function\\
$I^p_H(\chi_0)$ \> hypervolume of the set of points $\chi_0$ relative to the reference point $p$\\
$\eta(\chi_0)$ \> nadir point\\
$\iota(\chi_0)$ \> ideal point\\
$\bar{\eta}$ \> conservative nadir point\\

\textit{Flapping wing aircraft example}\\

$DI$ \> wing dihedral\\
$TW_i$\> internal twist\\
$TW_e$\> external twist\\
$a_{DI}$ \> amplitude of the wing dihedral\\
$p_{DI}$ \> period of the wing dihedral\\
$a_{TW_i}$ \> amplitude of the internal twist\\
$p_{TW_i}$ \> phase of the internal twist\\
$r_{TW_i}$ \> reference of the internal twist\\
$a_{TW_e}$ \> amplitude of the external twist\\
$p_{TW_e}$ \> phase of the external twist\\
$r_{TW_e}$ \> reference of the external twist\\
$\tau$ \> instantaneous torque\\
$\omega$ \> instantaneous rotational speed\\
$P$ \> mechanical power\\
$b$ \> aircraft wingspan\\
$S$\> wing area\\
$c_m$ \> mean chord\\
$U$ \> aircraft cruise velocity\\
$\nu$ \> air kinematical viscosity\\
$Re$ \> Reynolds number\\
$St$ \> Strouhal number\\
$k$ \> reduced frequency\\
$\kappa_1$ \> reduced internal twist frequency\\
$\kappa_2$ \> reduced external twist frequency\\

\textit{Basal ganglia model}\\

$BG$ \>basal ganglia brain nuclei\\
$CBG$ \> Contracting Basal Ganglia model \cite{girard08}\\
$GPR$ \> Gurney, Prescott and Redgrave basal ganglia model \cite{gurney01}\\
$WTA$ \> Winner-Takes-All\\
$GPi_j$ \> $j$-th channel of the basal ganglia output (Globus pallidus internal)\\
$sc$ \> selected channel\\
$nsc$ \> non selected channel\\
$GPe$ \> Globus Pallidus external\\
$D1$\> Striatum neurons with D1 dopamine receptors\\
$D2$\> Striatum neurons with D2 dopamine receptors\\
$FSI$ \> Fast Spiking Interneurons\\

\end{tabbing}

\section*{Introduction}

%


%

%



Computational models are now pervasive in most domains of science, including
astrophysics, mechanics, neuroscience, geophysics, to name a few.
They allow to compute the behavior of a system out of a particular context and of specific mathematical relations, thus allowing the simulation of the observed phenomenon \cite{Humphreys2002}. 
The specificity of the scientific approach relying on such models is that the human researcher generally doesn't master the whole process ranging from the initial mathematical model to the behavior exhibited in a particular context, leading to a so called {\it epistemic opacity} \cite{Humphreys2009}:
\begin{quote}
{\it
A process is epistemically opaque relative to a cognitive agent X at time t just in case X does not know at t all of the epistemically relevant elements of the process.
}
\end{quote}

Such an opacity stems from the fact that scientific work in this context involves so complex computations, that no human or group of humans can handle it without the help of a computer, thus leading to an hybrid scenario in which humans and machines are both critical. Likewise, the required validation of a computational model involves fitting model outputs with experimental data. A trial and error process may then be required to find the most adapted parameters among those that are justified by the assumptions made while building the model. This intrusion of heuristics, although perfectly justified \cite{Humphreys1995}, also adds some opacity in the research process based on computational models.

A computational model allows to construct new knowledge on the basis of what is learned during the model construction process and of the manipulations it allows \cite{morgan1999}. \cite{Hughes1997} proposes to consider the following steps in the process of knowledge acquisition from a model:
\begin{itemize}
\item \textit{denotation} step: relations are built between the model and the target, e.g. model subparts are justified or discussed on the basis of current knowledge;
\item \textit{demonstration} step: investigations on the features of the model; 
\item \textit{interpretation} step: conversion of the findings from the model to the target system.
\end{itemize}
On a complex model, the demonstration step may be particularly difficult because of the epistemic opacity. Features to study may naturally stem from the construction process, but some properties -- as those emerging from the interaction of different parts, for instance -- can hardly be induced if no evidence drives the exploration in this particular direction. In this perspective, the computational model is in itself a complex system to study in order to unravel its unforeseen features. Such an object could be studied in the same way as natural systems, i.e. by building a simpler model exhibiting the same behavior, but there is a specificity that can be exploited to propose an alternative procedure: the ability to compute the behavior of the model allows to perform a huge number of experiments with different contexts or with different model parameters. This particular feature can be exploited in a systematic way to generate experimental data that will help studying and then understanding such models. 

%
We propose such a method to help analyzing computational models thanks to a procedure generating dedicated experimental data. Such data are not outputs of the model, but sets of parameter values of the model associated to selected behaviors. The method automatically finds both parameter sets and behaviors, thus opening the way to the analysis of these particular behaviors. The method relies on the idea that the model optimizes more than one function. Such functions may correspond to an evaluation of how the models fits to several different experimental data or of how the model fits to other validated analytical models of the studied phenomenon. Once such functions have been defined, the method consists in using multi-objective optimization algorithms in order to generate a dense approximation of the set of trade-off solutions. Looking for a dense approximation and not a sparse one opens the way to the study of regularities in both parameter and behavior spaces.

To reach this goal, we propose to use multi-objective evolutionary algorithms, that are efficient optimization algorithms in a multi-objective context and that also feature a high versatility, allowing to optimize any kind of model for any kind of objective function. The only restriction results from the inner working of evolutionary algorithms: as they rely on the evaluation of up to millions of different solutions, computational models will need to be fast enough -- or available computational power needs to be high enough -- to compute their behavior a huge number of times.  The set of trade-off solutions will then be analyzed in order to gain new knowledge about the model,  like the importance of a particular parameter, the different possible model behaviors and their features, etc. 
The proposed methodology relies on the following steps:
\begin{enumerate}
\item defining the relevant functions to be optimized (at least two);
\item finding a dense approximation of the set of trade-off solutions;
\item analyzing these points both in the parameter space and in the space of optimized functions values, i.e. the objective space.
\end{enumerate} 
 
After a review of related work, some basic notions required to understand the different aspects of the methodology are presented. The different steps of the method are then proposed together with a list of questions that the method might contribute to answer. Some tools to be used to extract knowledge from generated data relative to these questions are also reviewed. Two examples are then considered and studied with the proposed methodology:
 \begin{itemize}
 \item a flapping wing aircraft model;
 \item a model of the basal ganglia brain nuclei.
 \end{itemize}
After a short introduction to the corresponding models, the methodology is applied to their study and the knowledge thus gained is highlighted. 




\section*{Related work}

Optimization methods in general and evolutionary algorithms in particular are frequently used in the engineering field to increase the profitability of an industrial process or the efficiency of a system. Besides this use, whose economical consequences has driven the optimization field for years, such methods can be used to support scientists' work. 

The most typical use consists in tuning parameters of models relative to given performance functions. The performance in this case is generally the closeness to available experimental data. A second and more specific use consists in considering the case of a natural system which has been optimized in some way through its history. Using an optimization process and looking at the results can be used as a validation for hypotheses on the objectives subject to optimization, or on the search space. Such an approach is used in biology, for instance, with the hypothesis that animals or plants had to optimize their ability to propagate their genes because of natural selection. A last use consists in using optimization in a data mining context, in order to discover new knowledge from a huge set of experimental data or even, and this is what we propose here, to generate data to dig in. 

\subsection*{Model  tuning}

Computational models aim at reproducing observed phenomena to validate hypotheses on the inner working of the involved systems and make predictions on the behavior to be observed in a different context. As the behavior of most models is highly dependent on some parameters whose values cannot always be deduced from direct observations, finding out the best parameter set to reproduce available experimental data is then a critical task for model designers. In this case, optimization is of particular interest with a cost function that typically measures the distance of model outputs to experimental data. Such use of optimization algorithms are frequent. 

In computational neuroscience, where the models are of increasing complexity, it has been used, for instance, to tune single neuron models \cite{VanGeit2008}, neural fields \cite{Igel2001,Quinton2010b} or neural network models of specific functions \cite{Cangelosi1997, Degris2004, Wang2007b}.

In hydrological modeling, model parameter identification is a difficult task because of their intercorrelations that may result in compensating errors. Optimization algorithms have then been used to identify model parameters and evolutionary algorithms revealed to be competitive with other optimization algorithms \cite{Duan1992, Wang1991}. Considering that antagonists objectives actually have to be optimized, multi-objective approaches have been proposed \cite{Ogou1998,gupta1998}, see \cite{Efstratiadis2010} for a review. Other physics related fields, like materials science, also consider model optimization with evolutionary algorithms \cite{Chaparro2008, Zhu2010,Aguir2011}, see \cite{Munoz-Rojas2010} for a review.

\subsection*{What is "optimized" by a particular process?}

Optimization tools can also be used to test hypotheses about a natural process that also performs some kind of optimization. In this case, a computational model is also optimized, but for a different purpose than mere model building. Several cost functions are defined and used to optimize the model. The one providing the closest behavior to the observed phenomena will be supposed to be the valid hypothesis.

Biologists use such an approach to study the criteria optimized by an animal. Starting with an idealized model that is generic enough, several cost functions are proposed and the resulting behaviors are compared to experimental data to see if they fit. The conclusions of such work may be the confirmation of hypothesis on the objectives definition and on the search space. Such an approach has been used to study biomechanical adaptations in bone tissue \cite{margerie2006} or migration behavior of birds \cite{Vrugt2007}, for instance. Likewise, hypotheses on the influence of wiring cost on neuron placement were investigated with the use of optimization algorithms \cite{Chen2006}.

\subsection*{Evolutionary algorithms and data mining}

Scientists have to face two different challenges, while studying a phenomenon:
\begin{itemize}
\item extracting new knowledge from experimental data in order to validate an hypothesis (relevance of a model, for instance);
\item and before that, generating relevant experimental data to analyze.
\end{itemize}

These two steps are of utmost importance in a scientist work. The first point consists in extracting an answer to a given question from a potentially huge set of data. This is the goal of the data mining field which has proposed algorithms to make this process more efficient, see \textit{Data Mining} section below for an introduction. Evolutionary algorithms have been proposed to support this step. In this context, evolutionary algorithms are used to induce rules that can predict future data \cite{Bongard2007Automated,Schmidt2009}, i.e. they are automatically building a predictive model. Other uses of evolutionary algorithms include clustering or data preparation \cite{Freitas2002}.

The second point corresponds to the design of the experiment that subsequent analyses will rely on. This step can hardly be completely automatized, but optimization algorithms can be part of it. Multi-objective evolutionary algorithms  revealed in different context to generate data from which new knowledge can be extracted. \cite{Efstratiadis2010}, for instance, used it to get some insight on model errors and uncertainties. \cite{Deb2006} used multi-objective optimization in an engineering context to generate data whose analyses helped discovering design principles for some devices. \cite{Deb}, for instance, used such an approach to find design principles for electric brushless motors. The approach, called \textit{INNOVIZATION} -- which stands for INNOVation through optimIZATION -- consists in first choosing some antagonistic objectives (two, at least), then using multi-objective optimization tools to get an approximation of the set of Pareto optimal solutions, and then look at the solutions to discover regularities specific to the considered system. Self-Organizing Maps (SOM) and Analysis of Variance (ANOVA) were also used to analyze designs resulting from a multi-objective approach and better understand the relations between design variables and objective functions for several aerodynamic design problems \cite{Jeong2005,Obayashi2007}. The proposed framework unifies such approaches and extends its use to different aspects of computational model analysis.

\section*{Background}

\subsection*{Multi-objective problems}

Optimization problems can be formalized as follows:

\begin{quote}
Find the parameter $\mathbf{X} =  
\begin{Bmatrix}
x_1 \\
x_2 \\
\vdots \\
x_m
\end{Bmatrix} \in \mathbb{R}^m  $ 
 that maximizes (or minimizes) $f(\mathbf{X})$
under the constraints:
\begin{eqnarray*}
 g_j(\mathbf{X}) &\leq& 0, ~~~j=1, 2, \dots, p\\
 l_j(\mathbf{X}) &=& 0, ~~~j=1, 2, \dots, q
\end{eqnarray*}
\end{quote}

where $g_j(\mathbf{X})$ and $l_j(\mathbf{X})$ are functions that express respectively inequality and equality contraints.

When $f(\mathbf{X})$ takes its values in $\mathbb{R}$, the problem is called a mono-objective problem. When $f(\mathbf{X})$ takes its values in $\mathbb{R}^n$ with $n>1$, the problem is called a multi-objective problem. A mono-objective problem generally has only a single optimal solution. Single objective optimizations result then generally in a single solution. In a multi-objective problem, when the objectives are antagonistic, multiple trade-off solutions do exist. Multi-objective optimization algorithms may then return either one particular trade-off or a set of different trade-offs. In this work, we will focus on this last case.

In a single objective context, comparing two solutions is straightforward: $\mathbf{X_1}$ is better than $\mathbf{X_2}$ if and only if $f(\mathbf{X_1})>f(\mathbf{X_2})$. In a multi-objective context, comparing two solutions is more difficult. A straightforward approach consists in computing a weighted sum of each component $f_i(.)$ of $f(.)$. With such an approach, $\mathbf{X_1}$ is better than $\mathbf{X_2}$ if and only if $\sum_{i=1}^n w_i f_i(\mathbf{X_1})> \sum_{i=1}^n w_i f_i(\mathbf{X_2})$. This approach consists in choosing \textit{a priori} the relative importance of each objective. Another approach is possible in which no such choice is required. Such an approach requires to define a new dominance relation adapted to the multi-objective case. In this work, we will use the Pareto dominance relation defined as follows:

\begin{quote}
  A solution $\mathbf{X_1}$ is said to dominate another solution $\mathbf{X_2}$, if
  both conditions 1 and 2 are true:
  \begin{enumerate}
  \item the solution $\mathbf{X_1}$ is not worse than $\mathbf{X_2}$ with
    respect to all objectives; 
   \item the solution $\mathbf{X_1}$ is strictly better than $\mathbf{X_2}$
    with respect to at least one objective.
  \end{enumerate}
\end{quote}

In the following, $\mathbf{X_2} \prec \mathbf{X_1}$ will indicate that $\mathbf{X_1}$ dominates $\mathbf{X_2}$, $\mathbf{X_2} \preceq \mathbf{X_1}$ will indicate that either $\mathbf{X_1}$ dominates $\mathbf{X_2}$ or that $\mathbf{X_1}=\mathbf{X_2}$. 

This dominance relation is not a strict but a partial ordering relation. The consequence is that, using this ordering relation, there are generally multiple optimal solutions. As we rely on the Pareto dominance relation, such solutions are called Pareto optimal solutions. The set of non-dominated solutions within the entire feasible search space is the called the globally Pareto-optimal set. 

The existence of multiple optimal solutions is the core of the proposed approach: a multi-objective optimization problem will be defined, its Pareto-optimal set will be searched for and analyzed using data mining techniques to provide new insights on the considered computational model.






\subsection*{Multi-objective optimization algorithms}

How to find the globally Pareto-optimal set? The models for which the proposed approach will be the most useful will be complex, non linear, discontinuous models, that may have a temporal aspect. Such models are the most difficult to understand and are then those for which the need for analysis tools is the most critical. This requirement implies to choose versatile optimization tools. Metaheuristics are then a good choice for their ability to deal with such problems \cite{Michalewicz1999}. They 
don't impose any mathematical constraint and are thus widely applicable, but with the drawback of a convergence to an \textit{approximation} of the optimal solutions. 

\subsubsection*{Multi-objective evolutionary algorithms}

Among metaheuristics, evolutionary algorithms have shown to be particularly efficient in a multiobjective context \cite{Deb2001}. 
This feature stems from a specificity of evolutionary algorithms: they optimize in parallel a set of solutions instead of a single one. It makes them easy to adapt to a multi-objective context as each point of the population may converge to a particular trade-off solution. Multiple algorithms have been defined -- like NSGA-II \cite{Deb2002} or $\epsilon$-MOEA \cite{Deb2003}, for instance -- and they proved to be efficient in multiple applications \cite{Coello2004b}.

Evolutionary algorithms rely on an inspiration from Darwinian natural selection and 
feature a set of candidate solutions, called a population. Each solution is evaluated and its performance is used to apply a selection pressure: fittest individuals survive to the next generation and generate proportionally more similar and new solutions than others. New solutions are copies of their ancestors with some random alterations, called mutations. They can also have several ancestors, inheriting some parts of each ancestor through a crossover operator. A typical evolutionary algorithm can be summarized as follows:
\begin{enumerate}
\item generate a random set of solutions
\item evaluate each of them
\item \label{it:eval} select the fittest solutions
\item generate new solutions
\item evaluate each of them
\item if the number of cycles is below a chosen threshold, go back to \ref{it:eval}, otherwise stop the optimization. 
\end{enumerate} 

See \cite{Eiben2008Introduction} for an introduction to evolutionary algorithms.

As evolutionary algorithms only require to evaluate the objective function $f(.)$ at different points, these algorithms indeed impose no mathematical constraints. They typically don't need to know the derivative of the function, for instance. Their drawback is that they require a huge number of evaluations to converge (typically from $10^5$ to $10^6$ evaluations). The consequence for the proposed approach is that the computational models under study must be fast enough to allow a huge number of computer simulations. 

\subsubsection*{Indicators}
\label{sec:indicators}

In the following, different indicators to be used within the proposed multi-objective analysis method are defined.

Let denote $\chi$ the set of approximations of the globally Pareto optimal set generated by a set of runs $\Gamma$. As evolutionary algorithms feature a stochastic behavior, each time an optimization is actually performed, a particular approximation of the globally Pareto-optimal set of points is also generated. $\chi_i$ will be the approximation of the Pareto optimal set generated by the $i$-th run.



The {\it attainment function} $\alpha_{\chi}(z)$ \cite{Fonseca2001} aims at measuring the performance of an algorithm while taking into account its stochastic feature. $\alpha_{\chi}(z)$ can be defined as the probability of finding at least one solution which attains $z$ (in our case that dominates $z$) out of the set of optimization results $\chi$. 
It can be estimated via its empirical counterpart defined over $r$ sets of optimization results as follows:

$$\alpha_r(z) = \frac{1}{r}\sum_{i=1}^r I\{\chi_i \unrhd z\}$$

where $I\{.\}$ is the indicator function, equal to $1$ if the assertion is true or $0$ else. $\unrhd$ represents the following relation:

$$ \chi_i \unrhd z~~\equiv~~\exists x \in \chi_i~|~x \succeq z$$

An {\it attainment surface} \cite{Fonseca1996} can then be defined as the hyper surface described by a particular value of the attainment function. Some particular attainment surfaces will be of particular interest in the following: the 1-attainment surface  and the 0-attainment surface. The 1-attainment surface $\Psi_1$ represents the set of the "worst" points among the solutions, i.e. the set of the most dominated points. The 0-attainment surface $\Psi_0$ represents the set of the "best" points among the solutions, i.e. the set of non-dominated solutions over all the results.

\begin{figure}[htbp]
\begin{center}
 \includegraphics[width=0.45\linewidth]{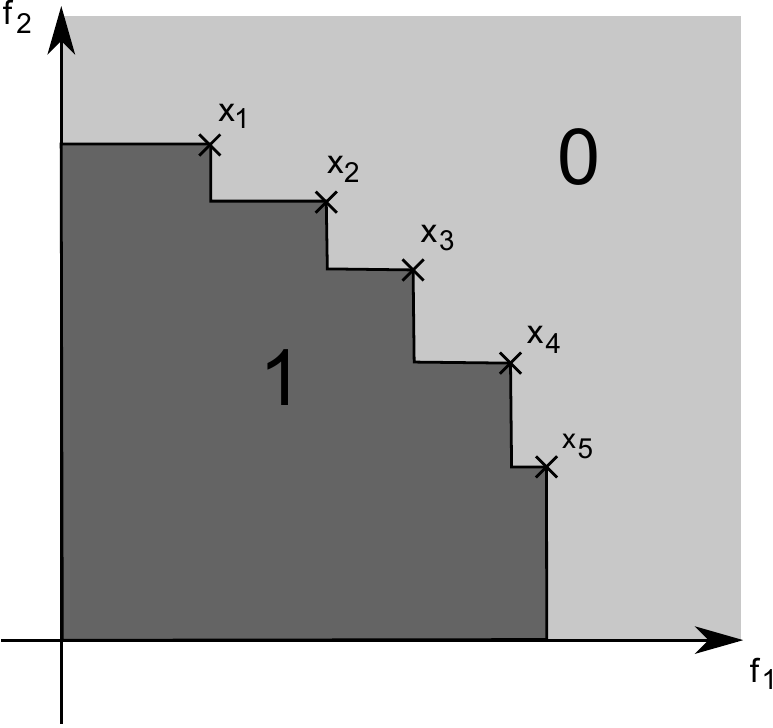}
~~~~~
 \includegraphics[width=0.45\linewidth]{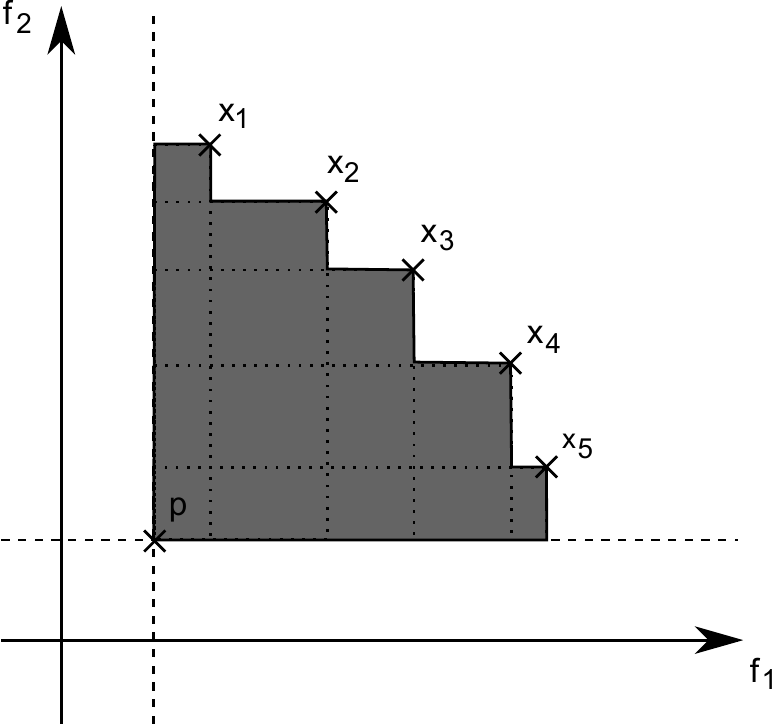}

\end{center}
  \caption{\label{fig:attainment_hypervolume}Left: values of the attainment function $\alpha_{\chi_0}(z)$ for $\chi_0=\{x_1, x_2, \dots, x_5\}$. Right: the hypervolume $I_H^p(\chi_0)$ is the area of the surface colored in gray.}
\end{figure}

In the following, as suggested in \cite{Zitzler2007b}, for the definition of the hypervolume described below,  we will simplify this definition and consider that $\chi$ is composed of only one set of points at a time, noted $\chi_0$. The definition of $\alpha_{\chi_0}(z)$ becomes (figure \ref{fig:attainment_hypervolume}, left):

$$\alpha_{\chi_0}(z) = \left\{ 
          \begin{array}{ll}
            1 & \qquad \mathrm{if}\quad \chi_0 \unrhd z \\
            0 & \qquad \mathrm{else} \\
          \end{array}
\right.$$

The hypervolume indicator of a particular non dominated set $\chi_0$, $I_H^p(\chi_0)$ \cite{Zitzler2003,Zitzler2007b} can be defined, for a given reference point $p$ as:

$$ I_H^p(\chi_0) =  \int_{\phi_p} \alpha_{\chi_0}(z) dz $$

with $\phi_p \subset \mathbb{R}^n$ the set of values above $p$, i.e. $\phi_p = \left\{ y \in \mathbb{R}^n | \forall i, y_i \geq p_i \right\}$. This is a generalized definition that holds for any dominance relation \cite{Zitzler2007b}. In the case of the Pareto dominance relation, as we will use here, and in two dimensions, it actually corresponds to the area of the union of the rectangles defined by the points $(p,x_i)$ with $x_i \in \chi_0$ (figure \ref{fig:attainment_hypervolume}, right).

The {\it nadir} objective vector $\eta(\chi_0)$ of a set $\chi_0 \subset \mathbb{R}^n$ is defined as the point with the least value possible for each objective:

$$ \eta(\chi_0) = \left\{ min_{y \in \chi_0} (y_i), i=1, 2, \dots, n\right\}$$

Likewise, the {\it ideal} objective vector $\iota(\chi_0)$ of a set $\chi_0 \subset \mathbb{R}^m$ is defined as the point with the maximum value possible for each objective:

$$ \iota(\chi_0) = \left\{ max_{y \in \chi_0} (y_i), i=1, 2, \dots, m\right\}$$

~\\

\subsection*{Data mining}
\label{sec:data_mining}

The ability to gather huge sets of data is facilitated by modern technologies. This has pushed forward the need to automate the process of extracting from such huge data sets consistent relationships between variables, leading to the Data Mining field. Using a combination of machine learning, statistical analysis, modeling techniques and database technology, data mining finds patterns and subtle relationships in data and infers rules that allow the prediction of future results. 

%
Two main families of data mining tools do exist: the first one, called the \textit{supervised data mining}, consists in identifying the relations between a \textit{target} variable $y$, (which can be a scalar, a vector or a tensor), and a set of $m$ \textit{predictors} $(x_1, x_2, \dots, x_m)$. The purpose of supervised data mining tools is to build out of available data a formal relation $y=F(x_{1}, x_{2}, \dots, x_{m})$ between the selected predictors which are identified as the most discriminant to explain and to model $y$.
The most famous techniques which belong to this family are:  oriented webs,  segmentation by decision trees and  neural networks.
The second family of data mining tools is called the \textit{non supervised} one. Its purpose is to characterize, inside a given population, homogeneous subgroups called clusters without any \textit{a priori }on a potential target variable. 
The famous techniques that one can find in this family are: non oriented webs, topology by Kohonen maps, two steps cluster, etc. 

The process of data mining actually consists of three steps: 
\begin{enumerate}
\item data processing. 
\item data modeling. 
\item deployment. 
\end{enumerate}

\textit{Step 1: data processing}. This stage usually starts with data preparation which may involve cleaning data, data transformations, selecting subsets of records. Then, depending on the nature of the problem, this first stage of the process of data mining may involve a simple choice of predictors for a regression model or need elaborate exploratory analyses using a wide variety of graphical and statistical methods in order to identify the most relevant variables that can be taken into account in the data modeling step. 

\textit{Step 2: data modeling}. This stage involves considering various models to identify the best one based on their predictive performance. It might seem like a simple operation but it sometimes involves a very complex and elaborate process. There are a lot of techniques developed to achieve that goal as we explain in the beginning of this section. The selection of the future operational model consists in applying the different potential models to the same data set and then comparing their performance to choose the best. 

\textit{Step 3: deployment}. That final step involves the model selected as the best in the previous stage to apply it on new data in order to generate predictions or estimates of the expected target. 



Numerous books review the theory and practice of data mining; the following references are  general books on data mining, representing a variety of approaches and perspectives \cite{Han2006,Hastie2001,Berry1999,Edelstein1999,Weiss1998}. 

\section*{Multi-objective analysis}

Multi-objective analysis relies on optimization tools, but its aim is not to find an optimal set of parameters. The method aims at helping the study of computational models by experts. Lots of computational models are an assembly of different parts whose behavior may be clearly known in separation, but whose interactions might be difficult to grasp. 
Multi-objective analysis method aims first at generating a set of specific solutions: Pareto optimal solutions, and then study their features. The goal of the method is then to find features of a model and identify points or regions of the parameter space that require further studies. In this context, the optimization tools are used not to increase the efficiency of a particular parameterized model, but to extract some knowledge on the considered model through the analysis of specific behaviors. The automatic selection of such behaviors is a key feature of the proposed approach. 

For a given parameterized model, the multi-objective analysis method aims at answering questions like:
\begin{itemize}
\item What are the objective values within reach?
\item What relations are there between the parameters and the objectives?
\item How critical is a particular parameter? Which parameter(s) make a solution Pareto optimal?
\item Are there a continuum or a set of qualitatively distinct solutions? In this case, where are the transitions and what are the features of the families of solutions?
\item What characterizes a particular behavior relative to the others?
\end{itemize}

Furthermore, such a method can also help comparing different models. Actually, when models can be evaluated on different criteria, one can be better than the other for some trade-offs, while the other may perform better in other parts of the Pareto optimal set. The comparison of Pareto optimal sets provides then a more rigorous view on the relative performance of the models. 

~\\

The method consists in performing the following steps:
\begin{enumerate}
\item Selection of the search space;
\item Definition of objective functions;
\item Computation of the Pareto front (or at least an approximation of it);
\item Analysis of the results.
\end{enumerate} 

The first three steps are classical to anyone familiar with optimization tools. The last step is the most important one of this methodology. We will list tools that can support it and highlight some of the questions they can answer. The examples provided below just aim at highlighting what a simple analysis can reveal on a computational model. A complete analysis of each of these examples goes far beyond the scope of this article. Following examples won't then exploit all of the analysis tools mentioned below.

\subsection*{Selecting the search space}

Choosing the search space is generally the most simple step. The model to be studied is known beforehand and generally described by a vector $\mathbf{X}$ of $m$ parameters: $x_1, x_2, \dots x_m$. The search space is then $\mathbb{R}^m$ if unbounded or a subset of it.

\subsection*{Defining objective functions}

Objective functions, that are also called cost functions or fitness functions, are the functions to be optimized. The most critical point at this step is to define at least two antagonistic objective functions. If the two functions are linearly dependent, a single solution can be optimal for both objectives: the set of Pareto optimal solutions will then be reduced to a single solution, as for a mono-objective optimization, thus not allowing to perform the proposed analysis.

Actually, any function numerically describing model features is worth considering for this step. This choice is clearly of critical importance for the analysis to be the most useful, as all the performed analysis will rely on these particular objective functions. The objective values within reach clearly depend on these functions, so do the relations between objectives and parameters and the other considered questions (how critical is a parameter? are there singularities? etc). Clearly, this step requires a deep expertise and must be carefully defined and justified for the results to be of any use. Typical functions to optimize are, for instance, accuracy of results and speed of response to both maximize or energy to minimize and accuracy to maximize, etc.

\subsection*{Computing the set of Pareto optimal solutions}

All further studies rely on the availability of the set of Pareto optimal solutions. In most cases, only an approximation of this set is within reach and the "true" set of Pareto optimal solutions is not known. It is then crucial to do whatever possible to guarantee the closeness of the approximated set to the real one. Furthermore, to get the best from the last step of the method, i.e. the analysis of the results, it is important to get a {\it dense} estimation of the Pareto optimal set in order to be confident in the fact that the analysis does not miss important features because of a too coarse sampling. It also opens the way to the study of regularities.

Multi-objective evolutionary algorithms (MOEA) are good tools to consider in this search as they can provide a dense estimation of the Pareto optimal set in a single optimization run \cite{Deb2001}. NSGA-II, for instance, is an efficient and yet simple algorithm that is now widely used in this context \cite{Deb2002}.  Evolutionary algorithms are stochastic optimization tools. Their estimation of the Pareto optimal set may then vary from one run to the other. Few theoretical work is focused on the study of MOEA convergence \cite{Hanne1999, Rudolph2000}. Most work in this field is based on empirical studies as theoretical results have often revealed to be of no practical use yet, as stated in \cite{Hanne1999}:
\begin{quote}
{\it
there is some indication that many multiobjective selection concepts used in practice may fail theoretically, it is possible that even such concepts without an almost sure convergence may work better in practice that those where (global) convergence can be proven.}
\end{quote}

For the INNOVIZATION approach, \cite{Deb2006} suggest to first use a multi-objective evolutionary algorithm, e.g. NSGA-II. The \textit{nadir} and ideal points are evaluated and a local search is used to get a better optimized front \cite{Benson1978, Deb2001}. Last, the normal constraint method (NCM) \cite{Messac2004} is used, starting at a few locations to check the accuracy of the optimized front. 

Considering that we only have access to approximations of the set of Pareto-optimal solutions and that they are generated by a stochastic process, we suggest another approach, aimed at exploiting at best all the available data:
\begin{enumerate}
\item repeat the optimization several times;
\item evaluate the disparity between the generated sets;
\item compute $\Psi_0$, the set of non-dominated solutions over all results;
\item for each point of $\Psi_0$, evaluate how close or how far it is to the closest points of other Pareto sets.
\end{enumerate}

Ideally, several different algorithms can also be used for the first step. Its goal is to increase the confidence in the results with regards to the stochasticity of the process used to generate them. The number of runs to actually perform should be high enough to ensure the statistical significance of the results, but the required computing power will limit it in practice.

The second step aims at evaluating the disparity between the runs. A too wide variability in the results is the symptom of a problem -- ill parameterized evolutionary algorithm, for instance : population size too small or too few generations -- thus suggesting that the generated solutions might be still improved. It suggests then to go back to the first step and perform other optimizations with new algorithms parameter values (or test other algorithms). A low variability increases the confidence in the fact that the runs have converged to a locally optimal set of solutions\footnote{It doesn't tell anything regarding the distance towards the global Pareto-optimal set.}.

The third step provides the best approximation of the set of Pareto optimal solutions relative to available data. Analysis of the shape and patterns of Pareto optimal solutions must then be performed directly on this set rather than on the result of a particular run. 

The last step aims at more precisely evaluating the disparity within the front, whereas the second step studied the global variability. A huge variability for a particular part of the front is a symptom that should be known by the expert performing the analysis. It is representative of parts of the search space that were more difficult to explore for whatever reasons (model instability, for instance) and it may be worth to further analyze these particular points to understand where the variability comes from, or it may also be worth knowing it in order to focus the analysis on points for which the convergence is less questionable. 

The first and third steps are straightforward. The fourth needs a dedicated evaluation algorithm that will be described below. The second  requires to evaluate a disparity between approximations of the set of Pareto-optimal solutions. In two or three dimensions, this can be evaluated visually by simply plotting the different fronts and looking at their differences. Making a quantitative comparison is not as simple as comparing results of single objective optimization algorithms. As it is required to compare multi-objective optimization algorithms, it has been the subject of intensive work over the last decade,  see \cite{Zitzler2008} for a review. Comparison of generated sets of solutions can be  performed on the basis of dominance ranking or performance indicators. 



Dominance ranking consists in evaluating if a particular set of solutions dominates another one, i.e. if the solutions of the considered set dominate each of the solutions of the other one. This method is aimed at ranking optimization algorithms, not at evaluating the disparity of the generated solutions and thus won't be used here.


Multiple performance indicators have been proposed to describe with a single scalar the performance of a multi-objective algorithm \cite{Knowles2002}. They are mainly aimed at comparing algorithms, but also at guiding the search or at defining a stopping criterion. To evaluate the disparity between the different runs, we suggest to use the {\it hypervolume indicator}, that has the drawback of an exponential computational cost relative to the number of objectives, but with the advantage of being strictly monotonic (see section \textit{Indicators} for its definition). 


To evaluate the disparity between runs, we propose to compute the following values:
\begin{itemize}
\item  For each of the performed run $r \in \Gamma$ , identify the nadir objective vector $\eta^r$ 
\item Discard the outliers: consider the conservative nadir point $\overline{\eta}$ defined as follows:
$$ \overline{\eta} = \left\{ max_{r \in \Gamma} (\eta_i^r), i=1, 2, \dots, m \right\}$$
\end{itemize}

The disparity between the generated sets is then evaluated as follows:
\begin{enumerate}
\item compute $\Psi_0$ and $\Psi_1$;
\item compute the difference of the hypervolumes  of $\Psi_0$ and $\Psi_1$ relative to the conservative nadir point: $I_H^{\overline{\eta}}(\Psi_0)$ and $I_H^{\overline{\eta}}(\Psi_1)$ to evaluate the variability of the runs.
\end{enumerate}

For the evaluation of the disparity within $\Psi_0$, we suggest to compute, for each point of $\Psi_0$, the distance to the closest points of each of the different Pareto sets $\Psi_r$ and keep the maximum value. This value may either be used visually or, with a distance threshold, used to define a conservative set $\Psi_{cons} \subset \Psi_0$ defined as the set of points $p\in \Psi_0$ for which the distance is below the chosen threshold.

To sum up this part, the evaluation of the Pareto front we suggest is the following:
\begin{enumerate}
\item repeat the optimization several times, if possible with several different algorithms
\item evaluate the disparity between the generated sets:
\begin{enumerate}
\item compute $\Psi_0$ and $\Psi_1$
\item compute their respective hypervolumes relative to $ \overline{\eta}$
\item if the relative difference is above a given threshold, go back to step 1 with different parameter values or with other optimization algorithms
\end{enumerate}
\item for each point of $\Psi_0$, evaluate the maximum distance towards each $\Psi_r$.
\end{enumerate}

\subsection*{Analysis of generated data}\label{Visualization_1}

What is available after the former step is an approximation of the Pareto-optimal set of solutions. Each point associates a particular set of parameter values $\mathbf{X}$ together with its corresponding objective values $f(\mathbf{X})$. Analyses proposed in the following aim at extracting knowledge from it. For each analysis, some typical questions the analysis might contribute to answer are listed.

\subsubsection*{Analysis of the Pareto set}~\newline

\textbf{Typical questions:}
\begin{itemize}
\item What are the objective values within reach?
\item What are the relations between objectives? 
\item Is the model ill-posed?
\end{itemize}

A first analysis focuses on the set of points in the objective space, i.e. in the space of $f(\mathbf{X})$ values. The lower and higher values for each objective as well as the shape of the set may be of interest for the information they provide on the computational model and on the set of contexts and parameters opened to the optimization. The shape of the front may highlight objective relations and an irregular shape may even be the symptom of an ill-posed model \cite{Efstratiadis2010}. 

Analyzing the Pareto set implies to draw at least a part of it. In two or three dimensions, as for the examples presented in the following, it can be directly plotted, but for higher dimensions, dedicated visualization tools have been developed, like the scatter plot matrix, the value path or the star coordinates methods \cite{Deb2001}. Subsequent analysis may rely on a direct observation of the relations between objectives, or on regression tools to identify it more formally.

\subsubsection*{Analysis of the parameters}\label{Visualization_2}~\newline

\textbf{Typical question:}
\begin{itemize}
\item What are the relations between model parameters and objectives?
\end{itemize}

Each point of the Pareto set corresponds to a particular set of model parameters. Looking at how these parameters change may highlight important features of the model, like parameter uncertainties \cite{gupta1998}. 
 
In the examples provided below, a particular parameter will be plotted against one or two objectives, thus not requiring complex visualization tools. Other analysis might need a global picture of the parameter set relation to objectives. Dedicated tools to project it in a two or three dimensional space are then required, like Kohonen maps \cite{Kohonen_2001}, for instance.

\subsubsection*{Analysis of solutions}~\newline

\textbf{Typical questions:}
\begin{itemize}
\item What are the typical behaviors of the model?
\item What are the features of a particular solution or of a cluster of solutions?
\item What are the most critical parameters?
\end{itemize}
 

As a Pareto set may contain a lot of different points and before undertaking deeper studies, it may be important to build clusters of solutions in order to identify the most salient behaviors. In some experiments, as in the second example presented below, the definition of such clusters may directly stem from expert knowledge. In other cases, automatic clustering methods can be used to detect groups of similar solutions on the Pareto front. The most commonly used algorithms to accomplish this task are k-means \cite{Zio_2010}, fuzzy c-means \cite{Zio_2010}, principal component analysis, hierarchical clustering \cite{Goel_2007} and Kohonen maps \cite{Hamdaoui_2010,Shinkyu_2005}. 

Once clusters of solutions have been identified, the proposed analysis consists in looking at which parameters better describe a cluster relative to other clusters. Understanding what makes a solution Pareto-optimal or what is specific of a particular cluster of solutions may highlight the relative importance of each parameter. To drive this analysis, classification algorithms can automatically extract the most descriptive cluster features \cite{Breiman}. The results of such classifications is a decision tree. Starting from the whole set of considered points, the tree aims at subdividing it according to variable ranges. Each internal node is associated to a variable and the branches coming out of it are labeled with ranges of values. Each path from the root node to a particular node is then descriptive of a set of points and the goal of decision tree algorithms is to find the description sets that discriminate best between two given set of points. When the study focuses on a particular cluster, it will highlight the main features of this cluster relative to other clusters (may it be from the Pareto set or from the set of explored points).

Further analyses may focus on a specific solution. While expert knowledge can be used to highlight a particular point, simple selection rules have also been proposed, like the one of  \cite{Blasco_2008}, that ranks Pareto optimal solutions by using a 
graphical method based on $L^{p}, \; p=1,2,\infty$ norms of the normalized vector of 
objectives. The chosen solutions are those that minimize the distance, measured in an $L^{p}$-norm, to the ideal point. The behavior of the chosen point can then be studied while relying on expert knowledge.

\section*{Example from Aerodynamics: study of a flapping wing aircraft}

The first example concerns the study of a flapping wing aircraft. Although the underlying physics is locally well understood, the impact of wing beats on aerodynamical forces and the question of what wing beats to generate to get a particular behavior of the aircraft is still a subject of intense research. The multi-objective analysis is used here to generate a set of wing beats to study. As few experimental data are available, the results will be related to what biologists did observe on birds of similar size. Biologists have identified relations between significant parameters like wing area, cruising speed, wing span, flapping wing frequency, wing load or mass \cite{tennekes96}. Wing kinematics have also been studied for several species \cite{tobalske96,park01,hedrick2002}. The goal of the multi-objective analysis is to identify some specific wing beats to be tested in an experimental device. The choice of the wing beats is driven by the need to enhance the model or more generally the understanding of the underlying physical phenomena.

\subsection*{Description of the model}

\begin{figure}
\begin{center}
\includegraphics[width=0.6\linewidth]{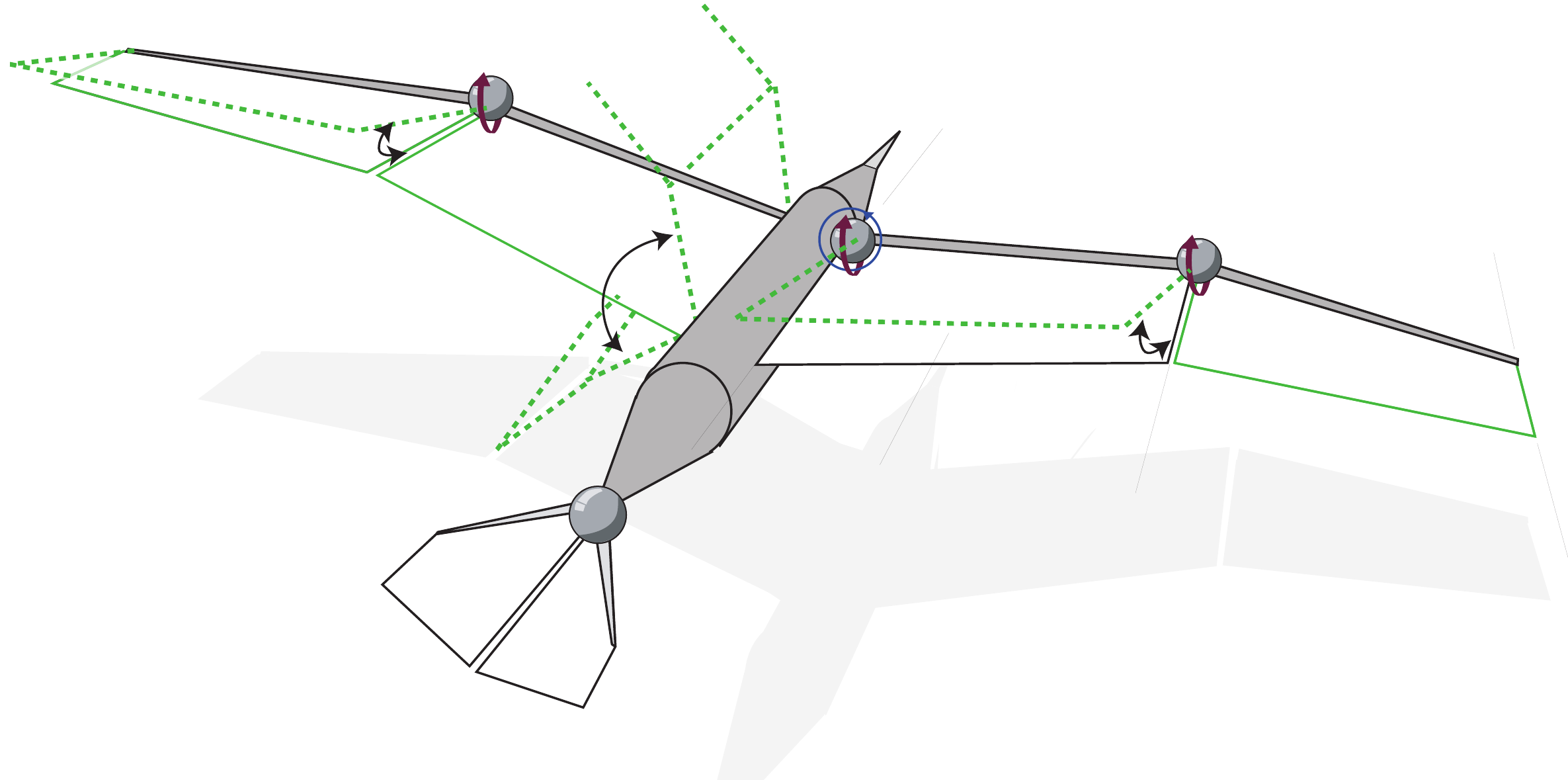}
\end{center}
\caption{\label{fig:oiseau-dim} Degrees-of-freedom of the simulated flapping wing aircraft.}
\end{figure}

Aerodynamical forces created by wing movements are computed with a semi-empiric, quasi-steady-aerodynamics model. Each wing is decomposed in three panels. For each panel, the local incident airspeed is evaluated. The computation of aerodynamic forces has been extended over classical lift computational models to better evaluate the forces at high incidence angles, see \cite{2008ACLI732,druot04} for a detailed description. The panels are considered as non deformable solids, connected to their neighbors via joints, as shown on figure \ref{fig:oiseau-dim}. The integration of the forces and the movements of the parts are computed using ODE software library\footnote{\url{http://ode.org}}.

Wing kinematics are described by the following equations:
\begin{eqnarray}
DI&=&a_{DI} \sin(2\pi t/p_{DI})\\
TWi&=&r_{TWi}+a_{TWi} \sin (2\pi(t/p_{DI}+p_{TWi}))\\
TWe&=&r_{TWe}+a_{TWe} \sin (2\pi(t/p_{DI}+p_{TWe}))
\end{eqnarray}

where $DI$ is the wing dihedral, $TWi$ the internal twist and $TWe$ the external twist. Wing kinematics are then described by eight parameters subject to the evolutionary optimization. The chosen ranges for each value are the following:
\begin{itemize}
\item amplitude of the dihedral: $a_{DI}$ in $[0;45^o]$
\item period of the dihedral (and of all other degrees of freedom): $p_{DI}$ in $[0.2;1s]$
\item reference of the internal twist: $r_{TWi}$ in $[-22.5 ; 22.5^o]$
\item amplitude of the internal twist: $a_{TWi}$ in $[0;45^o]$
\item phase of the internal twist: $p_{TWi}$ in $[0;1]$
\item reference of the external twist: $r_{TWe}$ in $[-22.5 ; 22.5^o]$
\item amplitude of the external twist: $a_{TWe}$ in $[0;45^o]$
\item phase of the external twist: $p_{TWe}$ in $[0;1]$
\end{itemize}

\subsection*{Objectives to optimize}

The energy-speed relation is characteristic of a particular shape and of a particular wing kinematic. We will then choose to optimize both speed and energy to empirically evaluate this relation. As we are interested in low as well as high speeds, we will perform two sets of experiments, one in which the speed is minimized and one in which the speed is maximized.


The objectives to maximize are the following:
\begin{itemize}
\item experiment 1:
\begin{itemize}
\item $+1\times$ average aircraft speed
\item $-1\times$ average mechanical power
\end{itemize}
\item experiment 2:
\begin{itemize}
\item $-1\times$ average aircraft speed
\item $-1\times$ average mechanical power
\end{itemize}
\end{itemize}

Following \cite{2008ACLI732}, the instantaneous mechanical power is computed as the scalar product between the instantaneous torque ($\tau$) and the instantaneous rotational speed ($\omega$) for a joint: $P=\sum_i | \tau_i . \omega_i |$. The mechanical power objective only takes into account the shoulder joints which are the main contributors to energy consumption. It should be noted that the power is always considered as positive. The mechanical power is then over estimated, as the torques required to accelerate or to slow down are considered as equivalent.

NSGA-II is used to perform the search, with a population size of 500 and during 1000 generations. Evolved parameters are represented as vectors of real values with a polynomial mutation and a sbx crossover, as described in \cite{Deb2001}, p124.

\subsection*{Results}

Four different runs have been performed for each of the two setups. 


\begin{figure}[htbp]
 \center
 \includegraphics[width=0.49\linewidth]{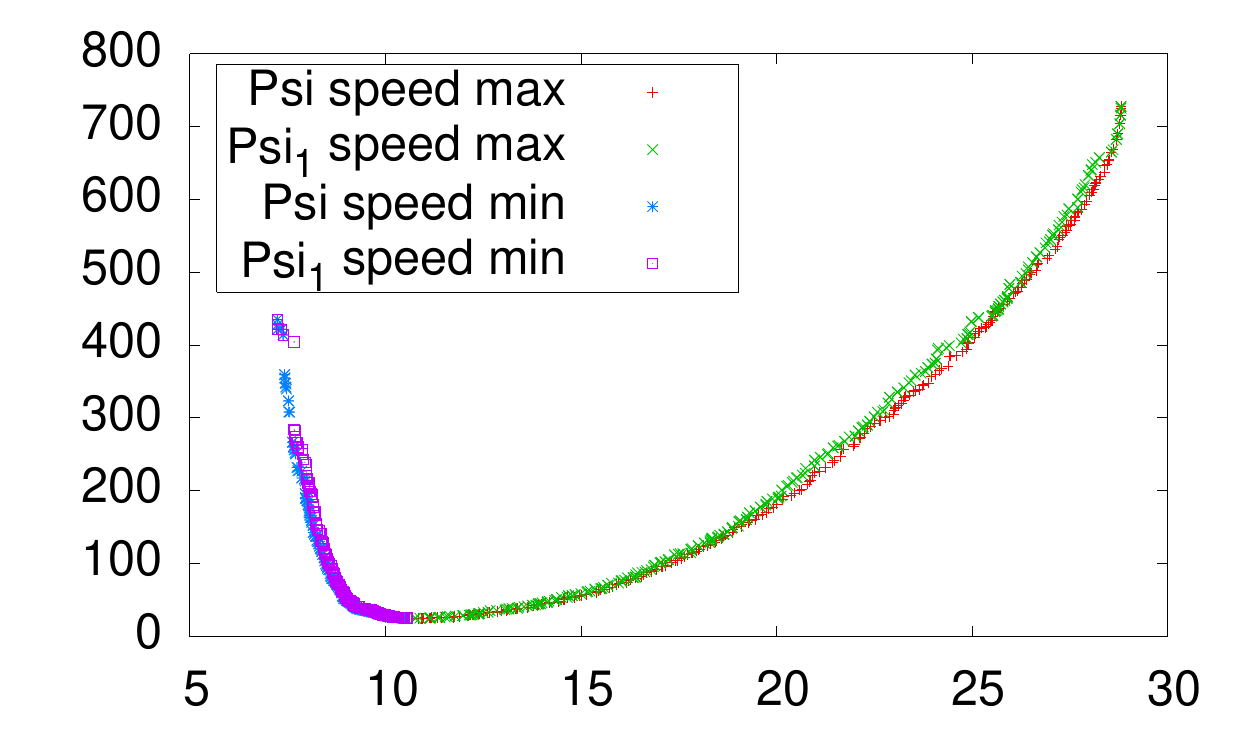}
  \includegraphics[width=0.49\linewidth]{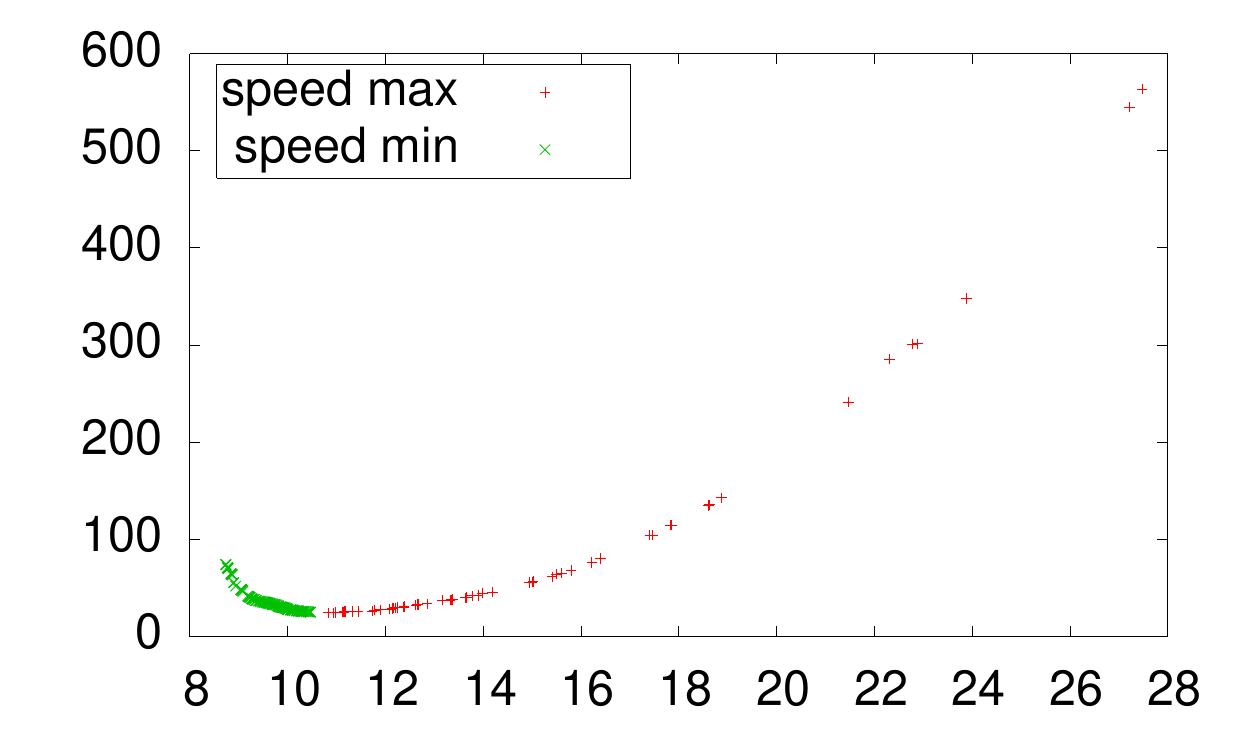}
  \caption{Left: $\Psi_0$ and $\Psi_1$ for the experiments on the flapping wing aircraft. "speed min" corresponds to the speed minimization experiments and "speed max" correspond to the speed maximization experiment. X-axis is the speed in $m.s^{-1}$, Y-axis is the energy in $W$. Right: Filtered version of $\Psi_0$ for the experiments on the flapping wing aircraft. Plotted points are those whose distance towards the Pareto fronts of other runs are lower than a chosen threshold $\epsilon$ (here $\epsilon=1$)}\label{fig:bird_psi}
\end{figure}

 
$\Psi_0$ and $\Psi_1$ have been computed and plotted on figure \ref{fig:bird_psi}, left. The conservative hypervolumes  are respectively $I^H_p(\Psi_0)=414.9$ and $I^H_p(\Psi_1)=393.4$ for the speed minimization experiment and $I^H_p(\Psi_0)=6267.4$ and $I^H_p(\Psi_1)=6117.3$ for the speed maximization. The difference is then $5.2\%$ of $I^H_p(\Psi_0)$ for speed minimization and  $2.4\%$ of $I^H_p(\Psi_0)$ for speed maximization. The disparity is then low, but not negligible, especially for the speed minimization case. The plot of the filtered version of $\Psi_0$ (figure \ref{fig:bird_psi}, right), highlights that  the differences are concentrated on the lowest speeds and, in a smaller extent, also on the highest speeds. This disparity reveals that the search was more difficult for these particular speeds. 

In the following, we consider the points from $\Psi_0$ and study their features from an aerodynamical point of view. The ornithopter is characterized by a wingspan $b$ of 1.93 m (both wings), a wing area $S$ of 0.407 m$^2$, a mean chord $c_m$ of $0.2$ and a variable cruise velocity $U$. The air kinematical viscosity $\nu$ is about $15\,10^{-6}$. To analyze the Pareto optimal solutions, the following aerodynamical dimensionless numbers are defined:
\begin{itemize}
\item Reynolds number: $Re =\frac{U c}{\nu}$
\item Strouhal number: $St = \frac{\sin(\frac{\pi \, a_{DI}}{180})b}{U \, p_{DI}}$
\item Reduced frequency: $k = \frac{\pi c}{U \, p_{DI}}$
\item Reduced internal twist frequency: $\kappa_{1} = 2 |\frac{\pi \, a_{TWi}}{180}| k$
\item Reduced external twist frequency: $\kappa_{2} = 2 |\frac{\pi \, a_{TWe}}{180}| k$
\end{itemize}

\begin{figure}[htbp]
\centering
\includegraphics[width=0.45\linewidth]{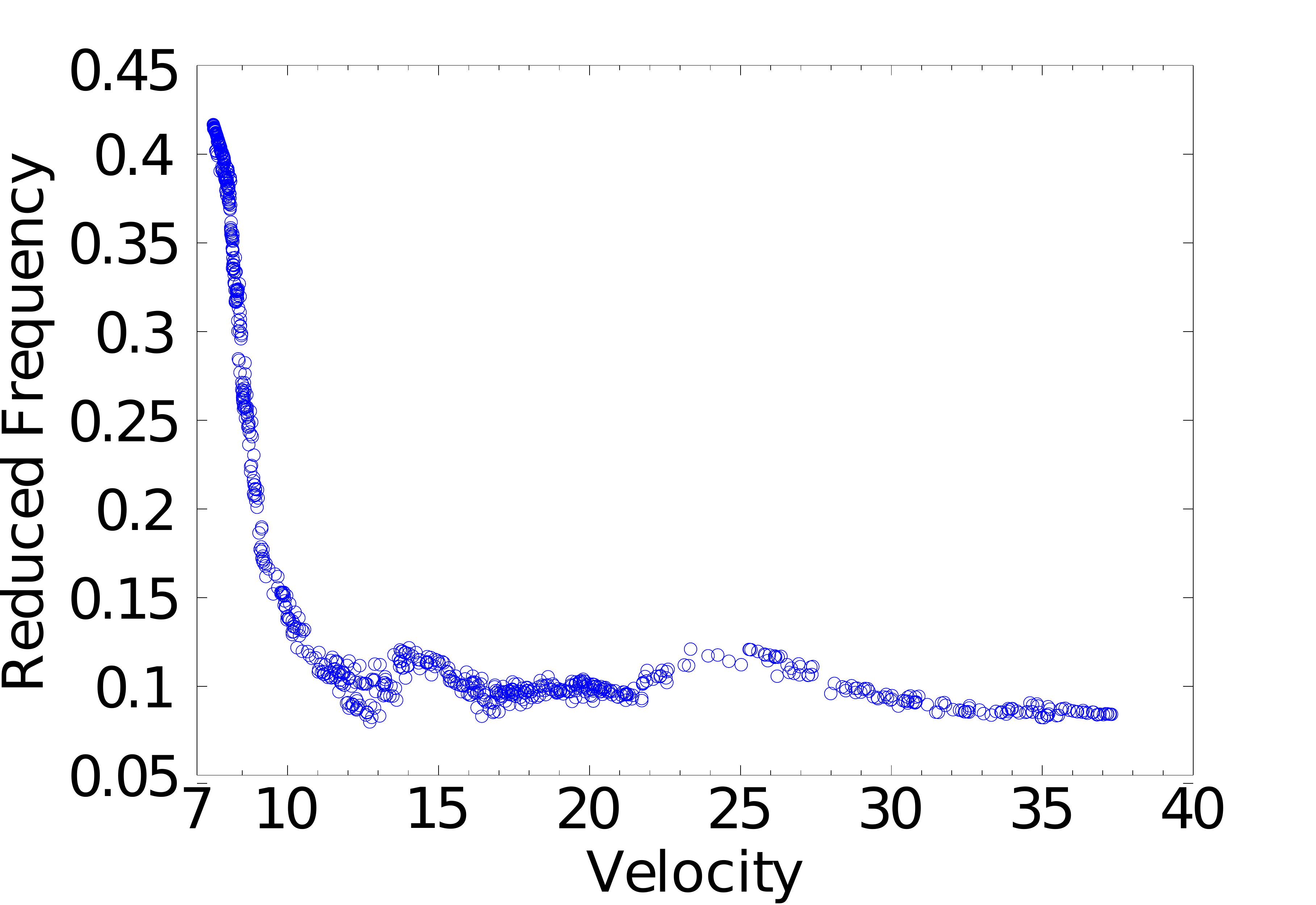}
~~~~~
\includegraphics[width=0.45\linewidth]{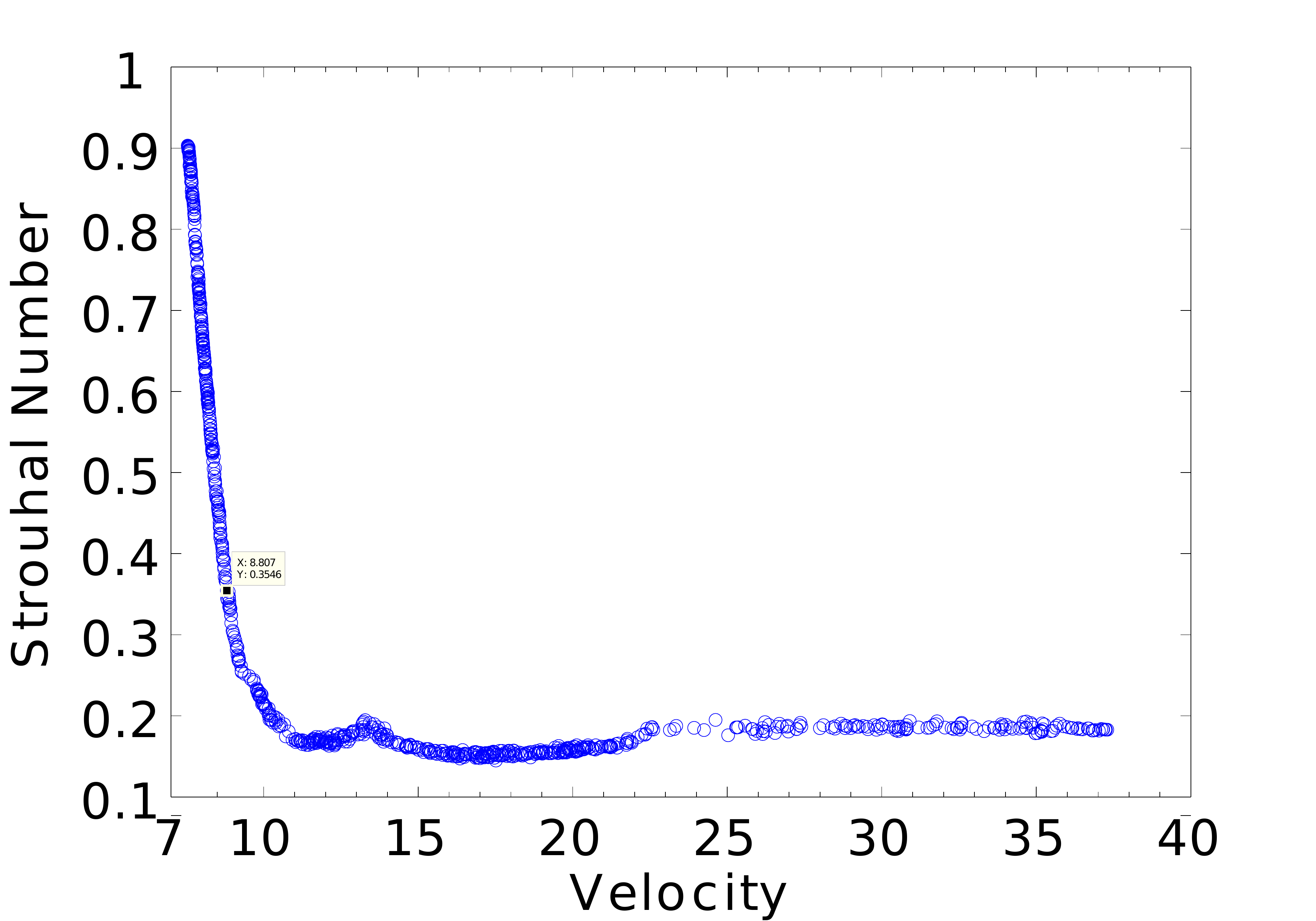}
\caption{Left: Reduced frequency versus the cruise Velocity. Right: Strouhal number versus the cruise Velocity}\label{k}
\end{figure}

If we look at the reduced frequency (see Figure \ref{k}, left), we can see that for velocities inferior to 8.7 m.s$^{-1}$ the reduced frequency varies between 0.25 and 0.42. These values are not standard for birds for which the reduced frequency oscillates between 0.1 and 0.25 \cite{Shyy_2008}. Thus, unusual high frequencies are experienced by our ornithopter at low speeds. These high frequencies may explain the optimization problems for low speeds, as they may result in an instability of the physics simulation. At higher speeds the reported reduced frequencies agree with the usual values encountered for birds.

Moreover, the Strouhal number (see Figure \ref{k}, right) oscillates between $0.35$ and $0.9$ when the cruise velocity varies from 7 to 8.8 m.s$^{-1}$, which is not common for birds (St $\sim$ 0.15-0.35) \cite{Taylor_2003}. This indicates that unusual high dihedral velocities are experienced by our ornithopter at low speeds. For higher velocities the Strouhal numbers agree with the usual values reported for birds.

\begin{figure}[htbp]
\centering
\includegraphics[width=0.45\linewidth]{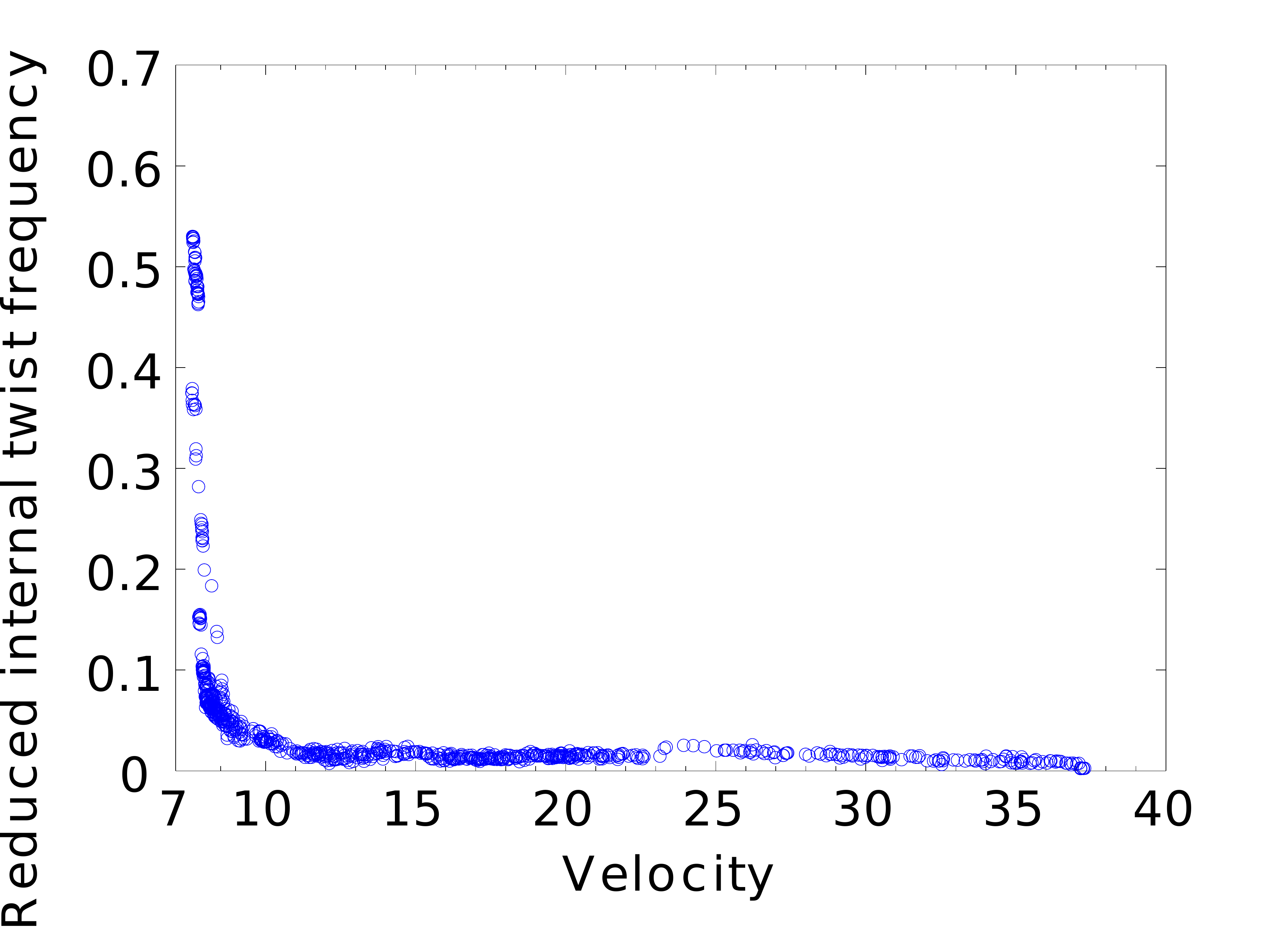}
~~~~
\includegraphics[width=0.45\linewidth]{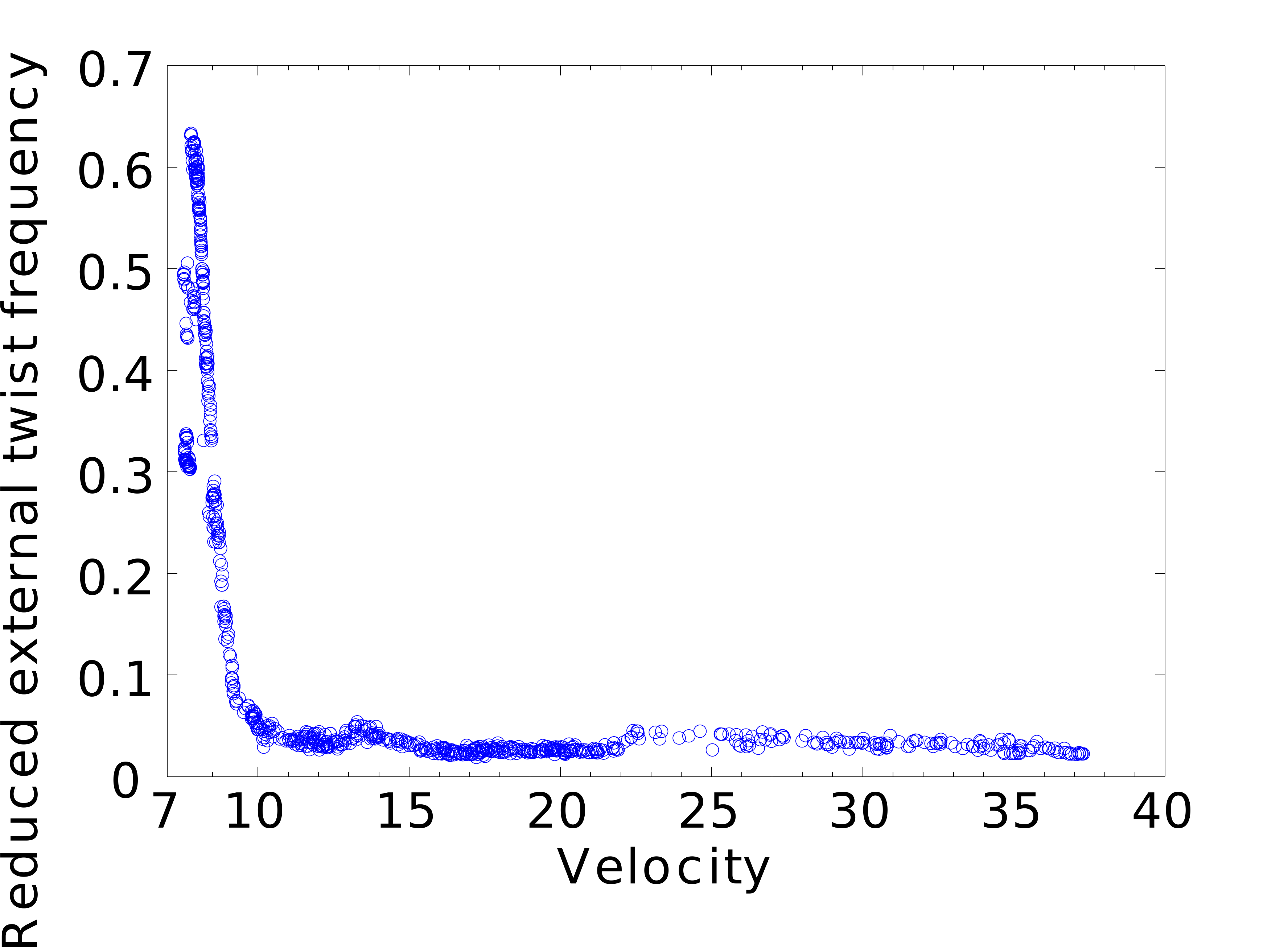}
\caption{Left: Reduced internal frequency versus the cruise Velocity. Right: Reduced external frequency versus the cruise Velocity}\label{ki1}
\end{figure}

In addition to that if we look at the reduced internal and external frequencies (see Figures \ref{ki1} left and right), we notice that they take high values at low speeds, the external reduced frequency $\kappa_2$ being slightly higher than the internal one $\kappa_1$. This means that our ornithopter experiences high twist velocities at low speeds.

\begin{figure}[htbp]
\centering
\includegraphics[width=12 cm]{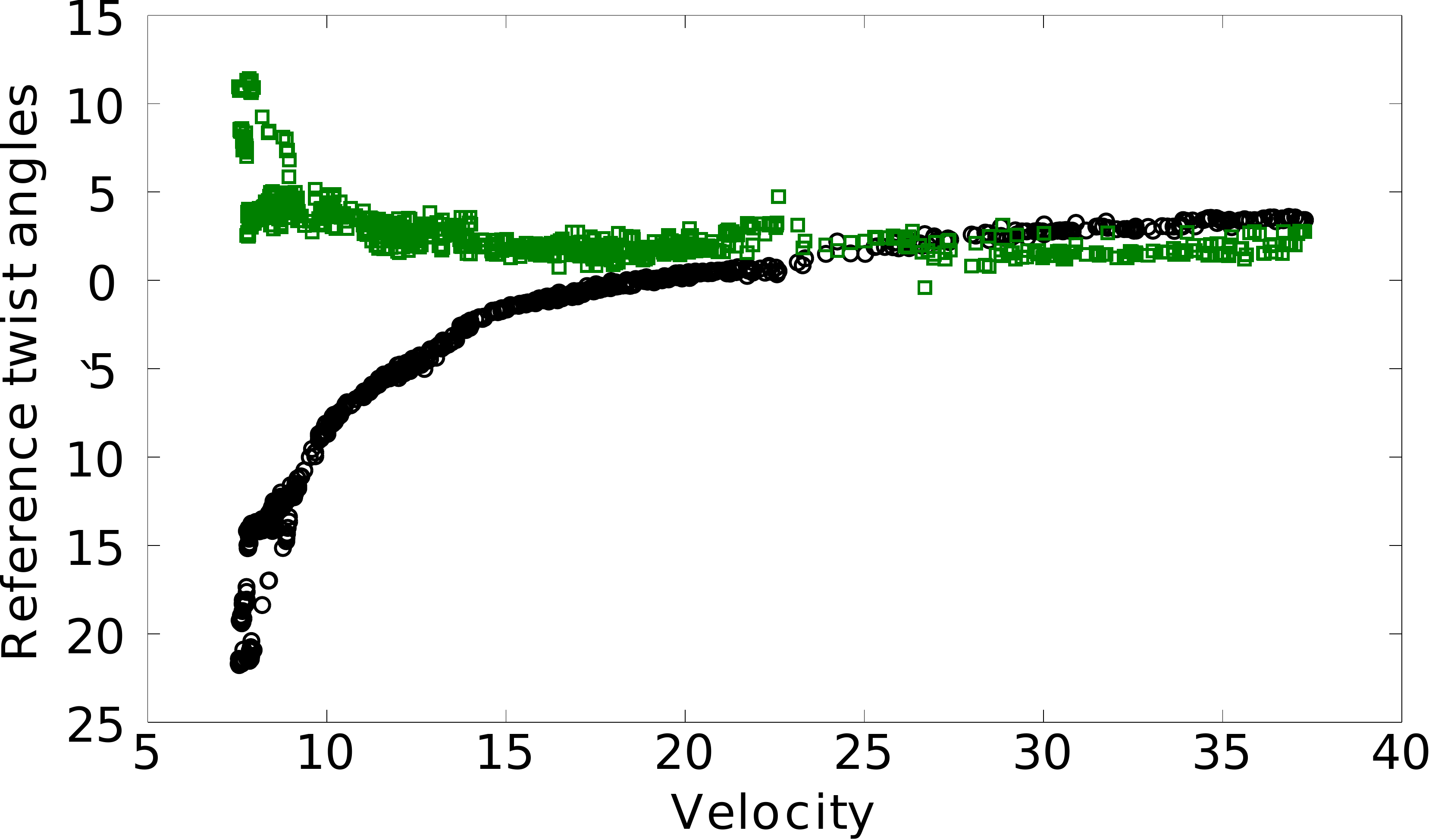}
\caption{Reference twist of internal ($\circ$) and external ($\square$) panels versus the cruise Velocity}\label{twist}
\end{figure}

Furthermore, if we take a look at $r_{TWi}$ and $r_{TWe}$ (see Figure \ref{twist}), we can see that the wing is highly twisted  at low speeds which is characteristic of the flight of birds at these speeds \cite{tobalske96}. At higher speeds the wing's twist decreases as expected \cite{tobalske96}.

To sum up the previous statements, we can say that low speeds are characterized by both high twist and dihedral velocities that can be attributed to the generation of the necessary lift at low speeds \cite{Walker_2002} while the high frequencies experienced are directly related to the generation of the power for flight \cite{park01}.

However, the reported values of Strouhal number and reduced frequencies for our ornithopter are quite high at these speeds. At this stage, to explain these values different hypotheses can be considered:
\begin{itemize}
\item the accuracy of the aerodynamical model \cite{2008ACLI732,druot04}  at low speeds is low;
\item the convergence of the optimization process at low speeds is too poor;
\item the ornithopter's configuration is not compatible with low speeds.
\end{itemize}

Concerning the aerodynamical model, if we compute the Reynolds number on the Pareto front we can see that it varies between $10^5$ and $5 \, 10^5$ while the velocity fluctuates between $7.5$ and $37.3$. At these Reynolds numbers, commonly experienced by birds similar to our ornithopter \cite{Spedding_2008}, transitional flows may occur \cite{Shyy_2008} especially at low speeds. Thus, one has to make sure that the aerodynamical model can handle with reasonable accuracy low Reynolds number effects by comparing the model with experimental results or CFD\footnote{Computational Fluid Dynamics} simulations. The kinematics associated to points in the low speed part of the Pareto front can then be used on a real experimental device in a wind tunnel or with CFD simulations to test the accuracy of the model and more precisely focus a study on the phenomena that the proposed model might neglect at this speed. 

%
%

The most interesting solutions from an energetic consumption point of view are those which minimize the energy. A characterization of these solutions versus the other solutions by a binary decision tree \cite{Hamdaoui_2010,Hamdaoui_2009} allows to identify the kinematic parameters that are important to define these interesting solutions.

A neighbourhood of the solution of minimal energy was defined by taking all the solutions such that the relative variation on the energy is less than 5 per cent. Then, a description of this neighbourhood relative to other Pareto optimal solutions was sought using Classification and Regression Tree algorithm (CART). It highlighted one significant rule:
\begin{equation}
0.48 \leq p_{DI} \mbox{ and } r_{TWi} \leq -5.85  \mbox{ and } 8.5 \leq a_{TWe} \label{r1}
\end{equation}

The obtained decision tree is 93 \% accurate and ranks the kinematical variables appearing in \eqref{r1} by order of importance. Thereby, $p_{DI}$ is more important than $r_{TWi}$ which is more important than $a_{TWe}$ to define the minimal energy solutions. This result is qualitatively acceptable from an aerodynamical point of view. Indeed the flapping period, the internal twist reference and the internal twist amplitude play an important role in the aerodynamics of our ornithopter \cite{tobalske96,Hamdaoui_2009}. Anyway, as there is no previous work on the subject, the only way to validate it is to launch the same experiments in a windtunnel and compare the obtained minimal energy solutions.

Furthermore, the precedent result gives us a compact vision of the kinematical model for the minimum energy solutions. For these solutions, the decision tree suggests that the important parameters for the model are the flapping period, the internal twist reference and the external twist amplitude. A possible way to check this result \textit{a posteriori} is to build a reduced model of the kinematics with just the three preceding kinematic parameters, recompute a Pareto front in the conditions stated before and compare the minimal energy solutions obtained. 

\subsection*{Summary}
We have optimized the parameters of the wing kinematics of a simulated flapping wing aircraft. To empirically generate its speed-energy relation, we used both speed and energy as objectives to optimize. Key steps and main results are listed below:
\begin{itemize}
\item At first, we evaluated the convergence of the optimization runs and highlighted a convergence problem in the low speed part;
\item we then performed an aerodynamical study of the Pareto optimal solutions. For speeds higher than $\sim$ 9 m.s$^{-1}$, these solutions are characterized by realistic values of Strouhal number, reduced frequency and Reynolds number which conforts us in the validity of these solutions and thus of the model at these speeds;
\item low speeds Pareto optimal solutions are characterized by highly twisted wings, high frequencies, high dihedral and twist velocities inherent to the flight at these speeds but the order of magnitude of the reported values is not realistic which shows that further investigations are needed. Three potential hypotheses may explain the high values encountered at low speeds: the inaccuracy of the aerodynamical model, the failure of the optimization process or the incompatibility of the ornithopter's geometry and/or kinematics with these speeds;
\item the minimum energy solutions are characterized by three kinematical parameters: the flapping period, the internal twist reference and the external twist amplitude. This result gives us a compact vision or reduced order model of the kinematical model for these solutions.
%
\end{itemize}

\section*{Example from Neuroscience: study of a Basal Ganglia model}

The Basal Ganglia (BG) is a set of interconnected brain nuclei. Located under the cortex, the BG is present in all the vertebrates, its structure remaining largely identical throughout the species. In order to gain a better understanding on the mecanisms governing it, various neuro-computational models have been proposed (for a review, see \cite{cohen09}). Among them are the recent CBG model \cite{girard08} and the more classical GPR model \cite{gurney01}; the parameters of both of them have been evolved with MOEA in a previous work \cite{lienard10}. We summarize here the results obtained and we push further the multi-objective analysis. This analysis will focus in particular on the relative importance of connections between nuclei for each model, and does provide a way to compare the two models.

\subsection*{Description of the models}

\begin{figure}[htbp]
 \center

 \includegraphics[width=0.45\linewidth]{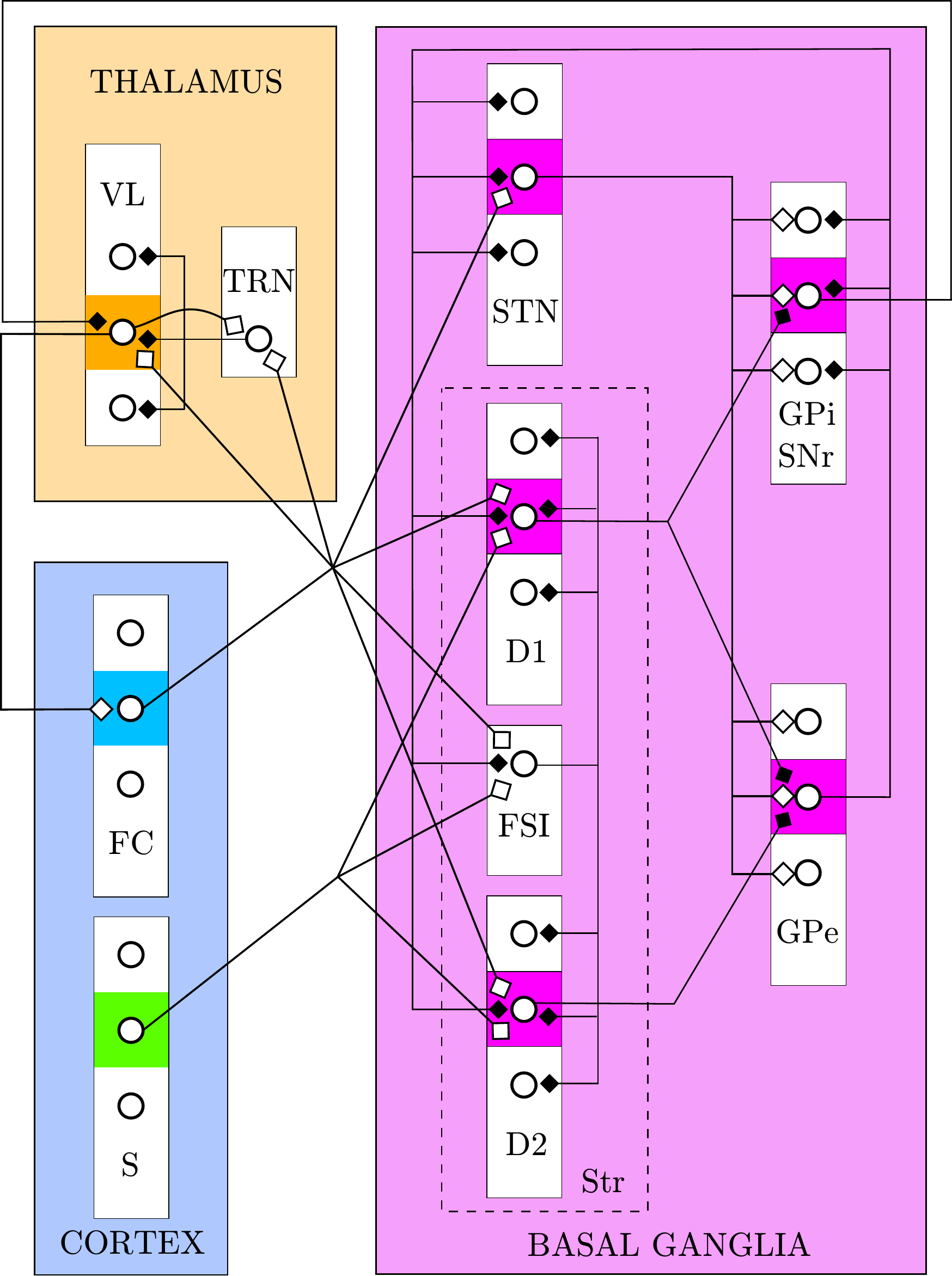}
 ~~~~~~
 \includegraphics[width=0.45\linewidth]{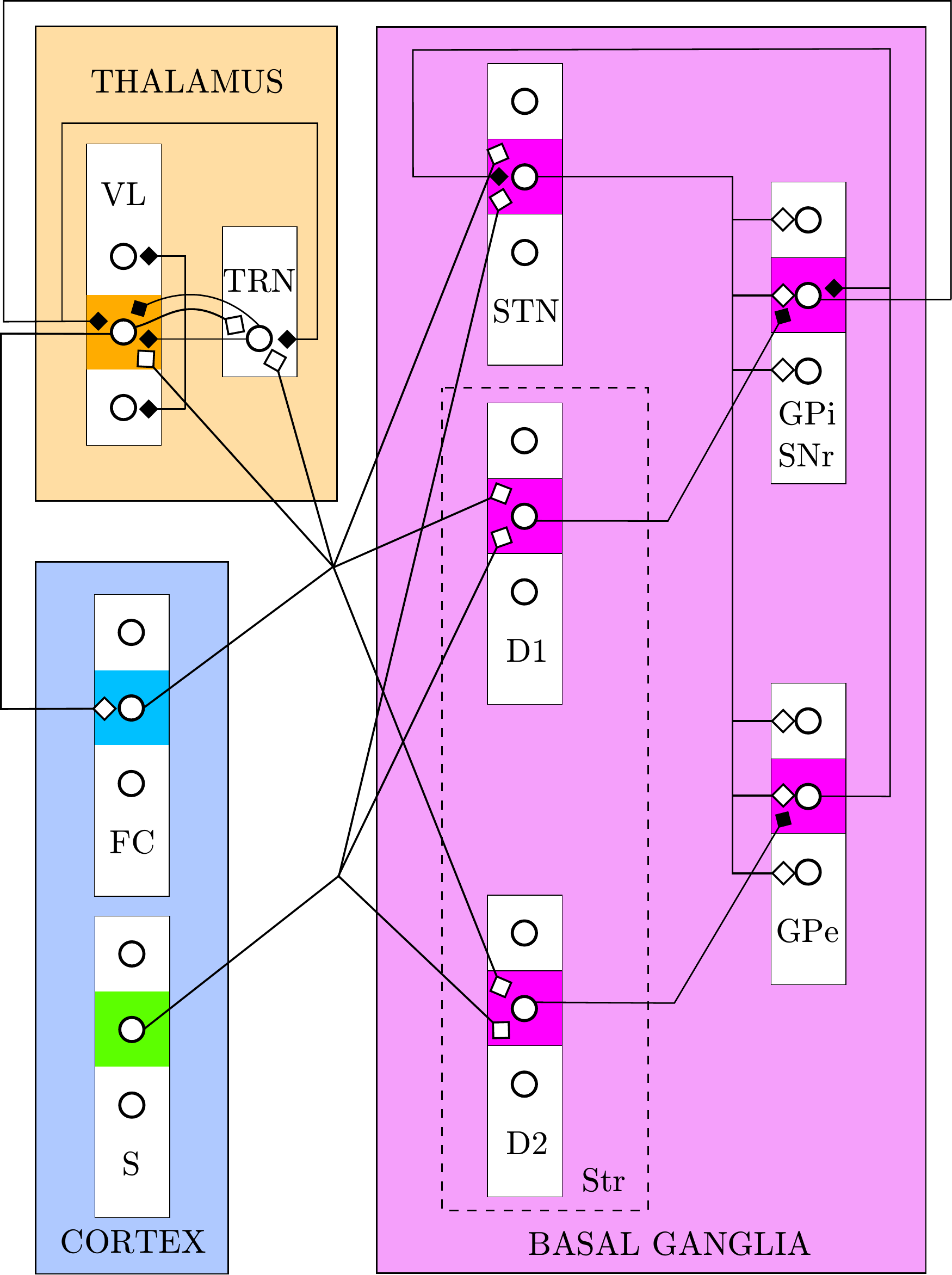}
 \newline

 \caption{\label{fig:basalganglia_architectures}
Schemes of the CBG (left) and GPR (right) architectures, representing nuclei from the Cortex, Thalamus and Basal Ganglia with labeled boxes . The inputs are the Saliences (S) in the cortex, and the outputs are the activity of the GPi/SNr neurons. Three channels in competition are represented in most nuclei. The outgoing connections of the shaded channel are the only one shown. White endings are excitatory, black are inhibitory.}
\end{figure}

The considered models of the BG are made of rate-coding artificial neurons (either leaky-integrators in the GPR or locally-projected dynamical systems in the CBG), representing the average activity of a population of real neurons. These models are basically described by a first order differential equation combined with a transfer function or a projection operator. They are represented by circles in figure~\ref{fig:basalganglia_architectures}. These neurons are grouped in nuclei which have been anatomically identified (boxes in figure~\ref{fig:basalganglia_architectures}). The main components of these models are the definition of the graph of connections between the nuclei, the weights of these connections and some internal properties for each nucleus. The architectures of the CBG and GPR models are respectively shown in the right and left panels of figure~\ref{fig:basalganglia_architectures}. The slight differences between them consist in the adjonction of connections in the CBG that were unknown or not included at the time when the GPR model was elaborated. The strengths of the connections and the internal nucleus properties account for 25 parameters for the CBG, and 20 for the GPR. Each connection parameter is to be evolved in the range $[0.05,1]$, as setting a connection weight to zero would be equivalent to delete this connection. For more details on the whereabouts of each models, we refer to the original articles \cite{gurney01,girard08} or to our previous work \cite{lienard10}.

The biological knowledge used to constrain the weights of the connections and the internal properties of the nuclei is rather weak, hence the modeler has to hand-tune almost every parameter to obtain a satisfying model. The MOEA were used to evolve the parameters while sticking to the architectures of the CBG and GPR models, in an attempt to shed light on the role of the various parameters and to compare the two architectures.

\subsection*{Objectives to optimize}

According to the mainstream hypothesis, the BG operates a generic selection system \cite{redgrave99}. The corresponding computational setup postulates the existence of a certain number of {\it channels} corresponding to possible alternatives; these are selected by disinhibition: the outputs of the channels are inhibitory and active by default, thus inhibiting their targets, when a channel is selected, its output is decreased, and thus its inhibitory effect canceled. The computational hypothesis is then formulated as a winner-takes-all among competitive channels, assigning the minimum activity to the channel with the highest input, while the others are activated as much as possible. This simple model of BG behavior will be used as a basis of the objective functions definition.

To ensure that the evolved models behave as a winner-takes-all, $N=500$ random input vectors have been submitted to each individual, drawn in $[0,1]$ from a uniform distribution. Two objectives were to be minimized. First, the channel corresponding to the largest input (the selected channel "sc") was to have minimal activity. Hence, the first fitness function is :
$$\displaystyle f_1 = \displaystyle\frac {\displaystyle\sum_{N} GPi_\mathrm{sc}} {N}$$

But this is not sufficient to obtain a WTA algorithm, as this could lead to the complete disinhibition of every channel. Therefore, a second objective was defined as the mean of all the other channels, the {\it not selected} channels "nsc", of cardinal "$\#(nsc)$" :
$$\displaystyle f_2 = 1 - \displaystyle\frac { \displaystyle\sum_{N} \left ( \frac{\sum  GPi_\mathrm{nsc}}{\#(nsc)} \right )} {N}$$

Any set of parameters minimizing these two objectives implements a WTA algorithm. Regardless of the inputs, a model with the fitnesses $(f_1=1,f_2=0)$ always selects all the channels, and a model with $(f_1=0,f_2=1)$ never selects any channel. All the other values for $(f_1,f_2)$ represent different approximations of a WTA.

\subsection*{Results}

\begin{figure}[htbp]
 \center \includegraphics[width=0.49\linewidth]{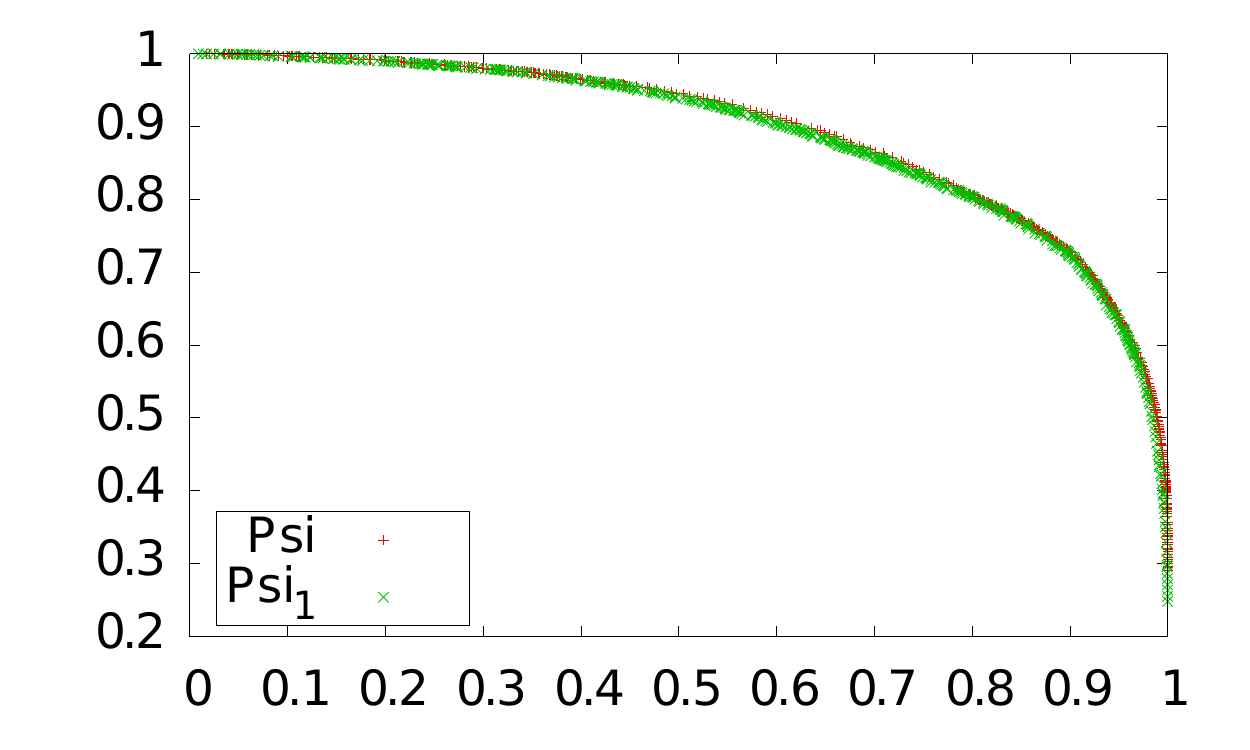} \hfill \includegraphics[width=0.49\linewidth]{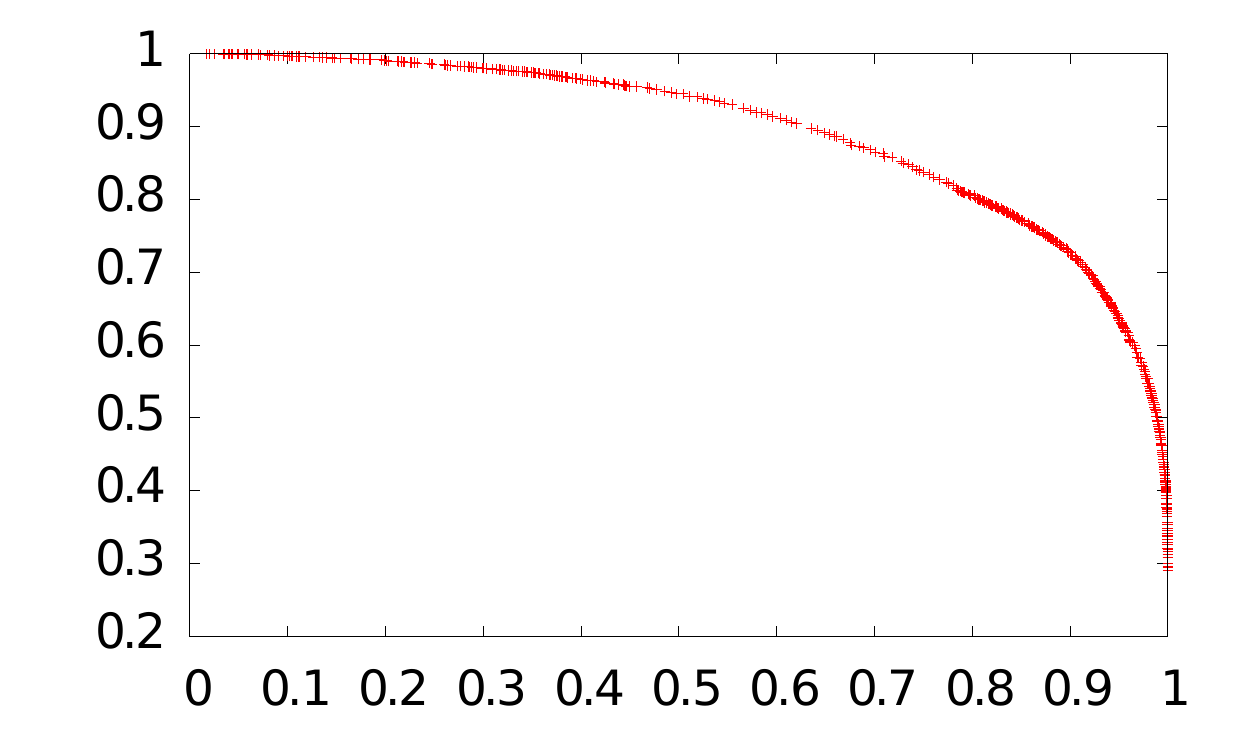}
  \caption{For both graphs, X and Y axis are the $f_1$ and $f_2$ fitnesses.\newline{\it Left: }$\Psi_0$ and $\Psi_1$ for the experiments on the basal ganglia model.\newline
  {\it Right: }Filtered version of $\Psi_0$  for the experiments on the basal ganglia model. Plotted points are those whose distance towards the Pareto fronts of other runs are lower than a chosen threshold $\epsilon$ (here $\epsilon=0.01$).
  }\label{fig:cbg_filtered_psi}\label{fig:cbg_psi}
\end{figure}

Ten runs have been first launched with a CBG architecture and their results compared and analyzed in details; ten runs with a GPR architecture were then computed for comparison purposes (see the comparison between the two architectures below). $\Psi_0$ and $\Psi_1$ have been computed and plotted on figure \ref{fig:cbg_psi}, left, for the CBG runs. Their conservative hypervolumes are respectively $I^H_p(\Psi_0)=0.593$ and $I^H_p(\Psi_1)=0.588$. The difference between the two is very low ($0.7\%$ of $I^H_p(\Psi_0)$). The confidence in the convergence of the runs is then high as the different runs have generated noticeably similar results. This is confirmed by the plot of figure \ref{fig:cbg_filtered_psi} (right) that shows a dense representation of the Pareto front with no particular gap, contrary to the previous experiment.


\subsubsection*{Interpretations of the entire set of solutions} Some connection weights are minimized or maximized for the whole Pareto front. This result can be interpreted as reflecting the relative contribution of the connections for the objectives. Hence, connections that are maximized at the upper bound of the search space, are of upmost importance for the selection. They reflect the parts of the circuit that are essential to achieve some kind of selection.

On the contrary, some connection parameters are minimized. Excluding the hypothesis that the mathematical formalism or the parameters search space are not adequate in modeling the reality, this means that either (1) these connections are indeed very weak, or even non-existent; or (2) these connections are useful for another purpose that is not expressed by the objectives.

\begin{figure}[htbp]
 \center\includegraphics[width=0.37\linewidth,angle=-90]{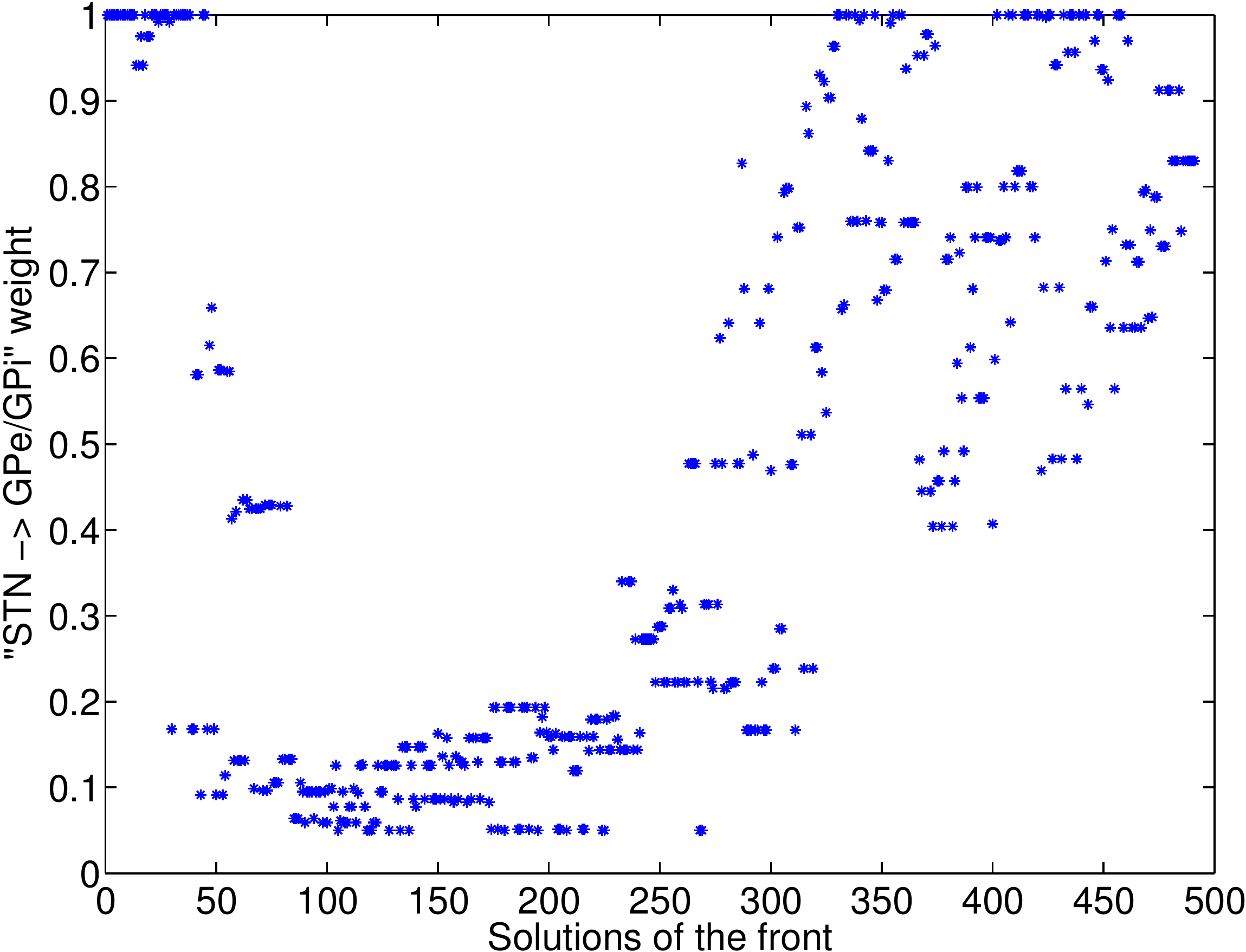} \hfill \includegraphics[width=0.37\linewidth,angle=-90]{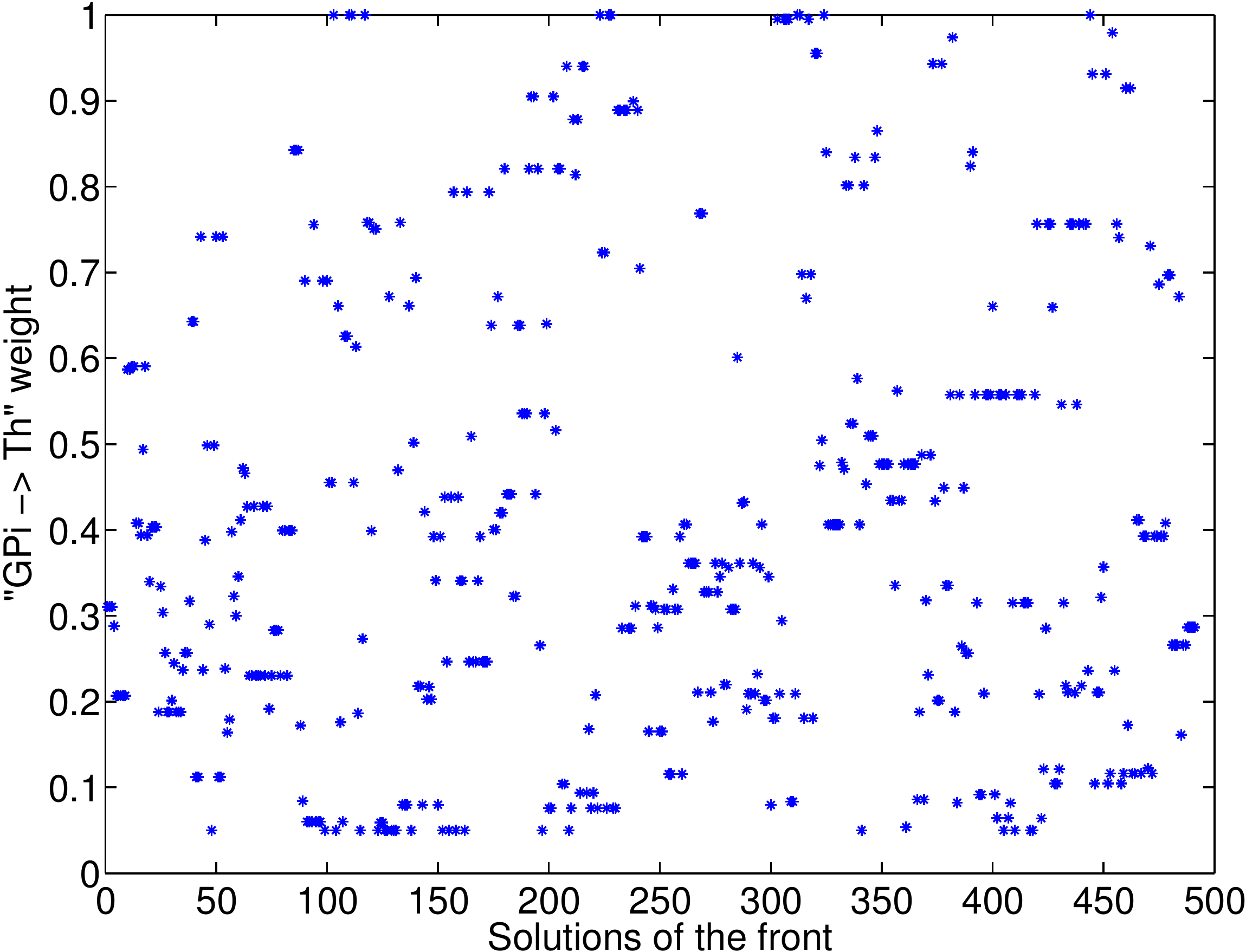}
 \caption{Plot of the STN $\rightarrow$ GPe/GPi (left) and the GPi $\rightarrow$ VL (right) connection weight parameters for each solution of the Pareto front. The X-Axis lists the solutions along the front, beginning with those who rank best for the first objective and ending with those who rank worst the first objective. While the STN $\rightarrow$ GPe/GPi appears to be set rather precisely by the evolution, the GPi $\rightarrow$ VL is apparently set at random.}\label{fig:basalganglia_params}
\end{figure}

Other parameters have been found to be set at random, this can be quantified with a very low measure of auto-correlation (see figure \ref{fig:basalganglia_params} for an example of a rather precisely set versus a randomly set parameter). The interpretation for these connections is that they are indifferent to the objective functions, neither contributing nor being against the supposed function.

We provide here only the guidelines for interpretations of parameters at the bounds or randomly set, as the biological implications for the specific parameters of the Basal Ganglia evolutions have been discussed in details elsewhere \cite{lienard10}.

\subsubsection*{Choice and analysis of a particular set of solutions} The whole set of solutions implements different kinds of selection, but some of these trade-offs are probably not interesting solutions. This is especially true for the solutions located at the endings of the pareto front, which tend either to never select any channel or to always select all the channels. Hence there is a need to choose a subset of solutions on the basis of expert knowledge. The comparative analysis of these solutions relative to the whole set of solutions aims to bring a characterization of the most interesting models.

\begin{figure}[htbp]
 \center
 \includegraphics[width=0.4\linewidth,angle=-90]{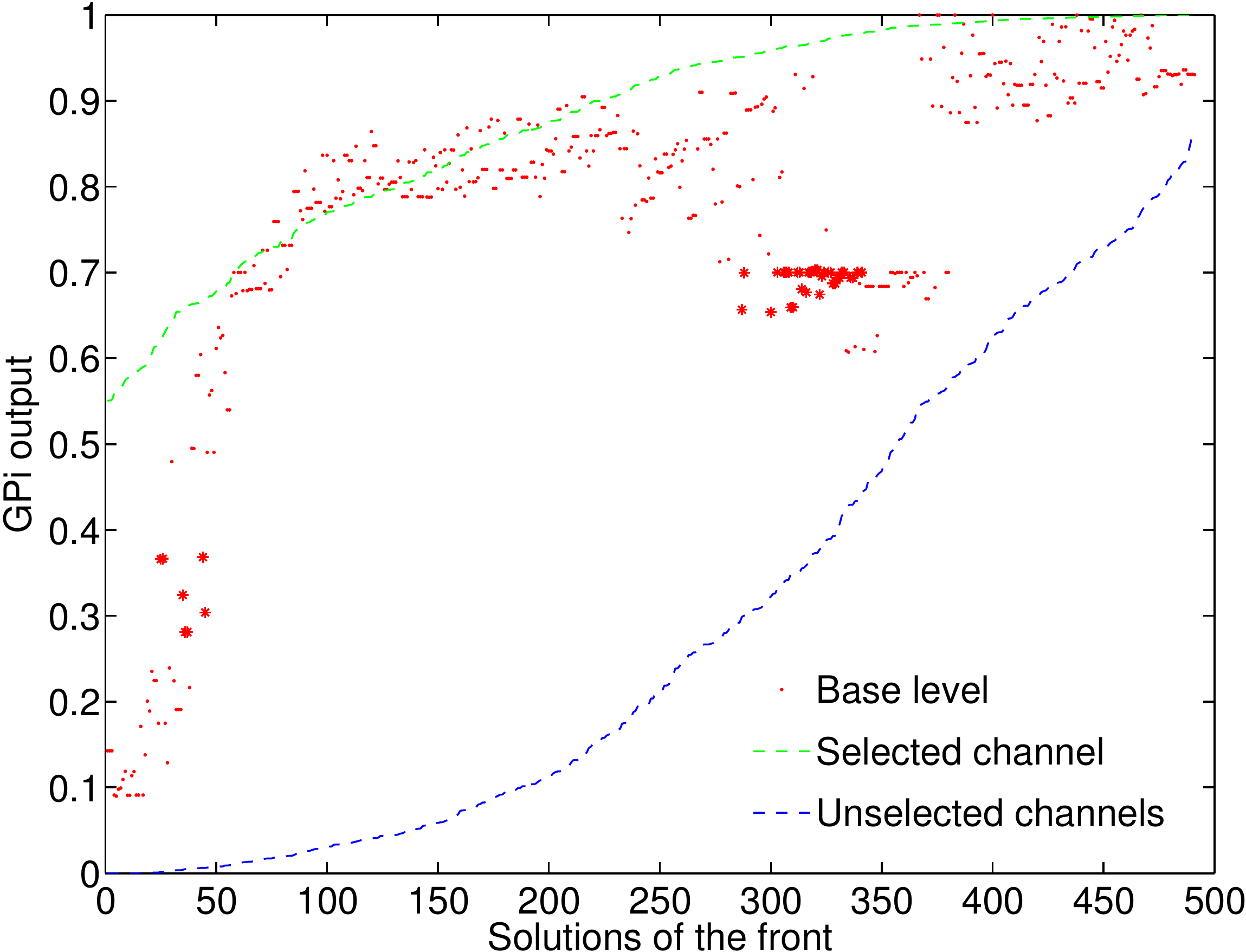}
 \caption{Plot of the output levels of each solution. The blue line corresponds to the mean output of the selected channel, and should be minimized according to objective 1. The green line corresponds to the mean outputs of the other channels, and should be maximized according to objective 2. The red dots correspond to the base level, i.e. the output in the absence of input. The dots corresponding to the most biologically plausible solutions (see text) are emphasized.}\label{fig:basalganglia_interestingzone}
\end{figure}

We began by evaluating the compatibility of the solutions with known biological data. To do this, we exploited two known facts. First, in the absence of input, the GPi, which is the output nucleus of the Basal Ganglia, is active. Its firing rate at rest, called the {\it base level}, should then correspond to a rather high value. Second, in a competition situation, the unselected channels should have an output higher than the base level, but the selected channel should have a lower output. By plotting the base level along the output of the selected channel as well as the mean output of the other channels, we have been able to identify a biologically plausible zone of the front (figure \ref{fig:basalganglia_interestingzone}).

\begin{figure}[htbp]
 \center
 \includegraphics[width=0.4\linewidth,angle=-90]{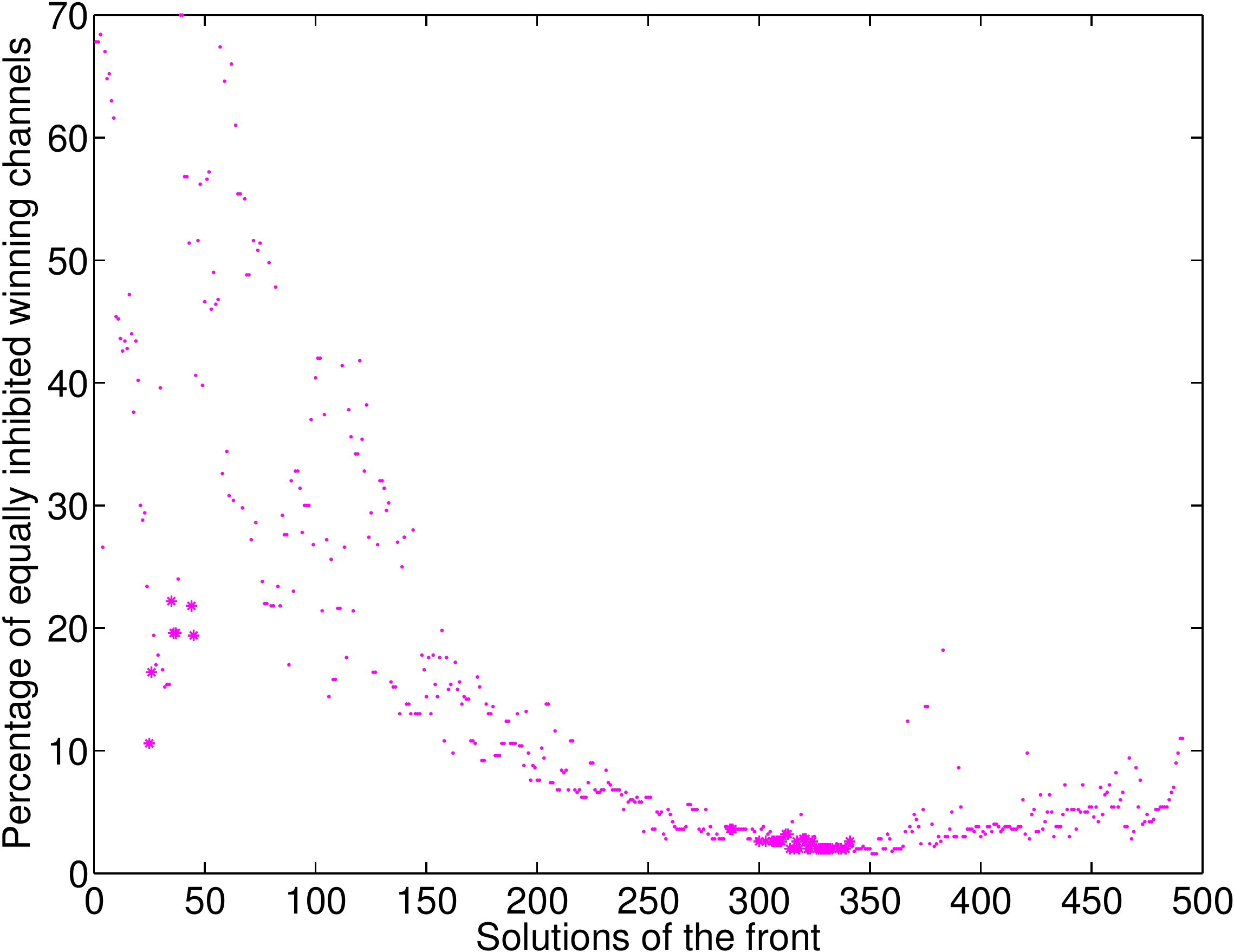}
 \caption{Plot of percentage of dual selection of the two best channels in input. The solutions labeled previously as "biologically plausible" are emphasized.}\label{fig:basalganglia_interestingzone_bis}
\end{figure}

Another way to evaluate the quality of the selection is with an engineering approach. We elaborated a measure of the ability to discriminate between the two best channels in input, by looking at how different their outputs are. Indeed, for many of the obtained solutions, when the channel with the highest input and its closest competitor are very close, both of them are selected simultaneously. To evaluate this, we scored the percentage of random inputs that led to the selection of two channels, i.e. that led to the case where the two best channels in outputs are closer than an arbitrary value of $0.01$. The results are shown on figure \ref{fig:basalganglia_interestingzone_bis}, they correspond roughly to the solutions we obtained with the previous criterion (figure \ref{fig:basalganglia_interestingzone}). This shows that the solutions operating a biologically plausible selection are also amongst the most efficient ones. 

\begin{figure}[htbp]
 \center
 \includegraphics[width=0.5\linewidth]{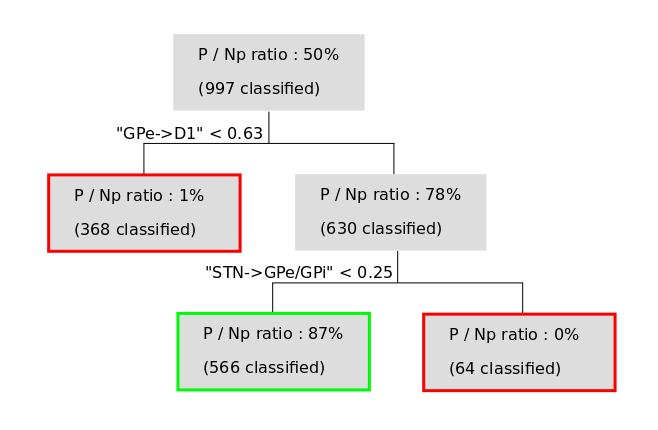}
 \caption{Binary decision tree representing the best parameters to discriminate between biologically plausible solutions ("P" solutions) and non-biologically plausible ones ("Np"). See text for methodology.}\label{fig:basalganglia_dt}
\end{figure}

By considering the subset of solutions efficient for both criterions, we can now analyze what makes them different from the other solutions, with the help of decision tree algorithms. To prepare this analysis, we began by normalizing the numbers of the solutions for each state by replicating the plausible solutions seven times, leading to a population comprising roughly as much plausible solutions (497) as non-plausible ones (501). After this, we applied a standard Classification And Regression Tree (CART) algorithm to separate the plausible "P" solutions from the non-plausible "NP" solutions (figure \ref{fig:basalganglia_dt}).

This decision tree shows that to ensure a biologically plausible selection, the GPe $\rightarrow$ D1 connection has to be superior than $0.63$. This is relevant to a biological interpretation, because the GPe $\rightarrow$ D1 connection is a subject of debate. Indeed, the connection from GPe to Striatum has been acknowledged for a long time in various species \cite{staines81,beckstead83}. Anatomically, \cite{bevan98} showed that some GPe neurons targeted preferentially Fast Spiking Interneurons (FSI), with approximately half of their axonal terminaisons ending on FSI and the remaining terminaisons ending on other neurons of the Striatum. Electrophysiologically, \cite{gage10} showed recently an inverse correlation in vivo between the GPe neurons firing rate and FSI firing rate, which is coherent with the idea that the GPe target them with an inhibitory influence. While these data suggest a strong influence of GPe neurons on FSI, they do not mean that their influence on the D1 and D2 neurons is weak or non-existant, and this connection has indeed been exploited in computational modeling work \cite{girard08}. The present result suggests that this connection should not only exist, but it should also be potent.

The percentage of solutions classified by the other branches of the tree is far less important, hence we have to be cautious in the interpretation of these. Furthermore, this would require a more detailed biological analysis, going out of the scope of this paper, so we choose not to continue further this analysis.

\begin{figure}[htbp]
 \center
 \includegraphics[width=0.5\linewidth]{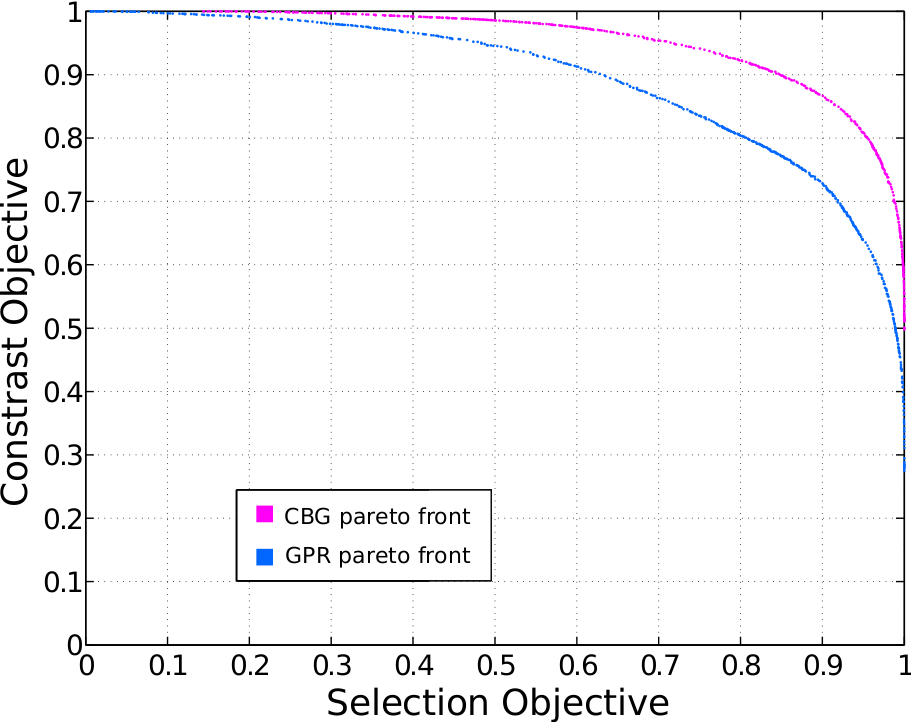}
 \caption{The fitnesses of the best fronts of the CBG (purple) and GPR (blue) fronts permit to compare the two architectures.}\label{fig:basalganglia_pareti}
\end{figure}

\label{sec:GPRcomp}\subsubsection*{Comparison between different architectures} The CBG model has been built upon another model, called the "GPR" model \cite{gurney01}. The main difference between the two models is that the CBG takes into account biological data that were unknown at the time of the elaboration of the GPR. This results in a slighty different connection scheme (figure \ref{fig:basalganglia_architectures}).

To evaluate whether the CBG or GPR architecture is more suitable to operate a selection process, the evolution of the GPR architecture has been done with a setup similar to the one done for the CBG. The parameters, which are fewer in the GPR (20 in place of 25), have been evolved under the same constraints. All the runs have converged toward the same approximation of the Pareto front, as they did for the CBG. Results showed that the solutions of the CBG strictly dominated the solutions of the GPR for the whole front (figure \ref{fig:basalganglia_pareti}). For any solution from the GPR front, there is a solution from the CBG front that has a better score for both objectives, resulting in a better selection as we defined it. This hence leads to the conclusion that the changes in connectivity, as well as the additional degree of freedom of the CBG, globally enhanced the selection functionality. This is particularly interesting as, to the extent of our knowledge, there is no other way to assess whether one architecture or another is more suitable to exhibit a selection process.

\subsection*{Summary}
We evolved the "CBG" architecture of the Basal Ganglia, to make it operate a selection amongst its entries. To this aim, we defined two antagonist objectives scoring the quality of the selection done on a given set of inputs. We list below the key steps and results of the multi-objective analysis of the evolution:
\begin{itemize}
  \item Before interpreting the results of the evolution, we made sure that they converged to a good approximation of the Pareto-optimal solutions.
  \item We analysed the whole set of optimal solutions. Some parameters are either maximized, minimized, or set of at random. We provided guidelines for interpreting the functional signification of these parameters.
  \item We then analyzed the specific set of interesting solutions, after having characterized it by two criterions. By using a binary regression tree, we are able to make a prediction concerning a specific connection, the GPe $\rightarrow$ D1 connection. This connection has a key role in the selection operated, and should be set at a rather high value. This result is particularly interesting as, even if the existence of this connection has been known for a long time, it has rarely been used in modeling works.
  \item Finally, we did optimize parameters of another architecture which is named "GPR", with the same setup as for the CBG. This showed that the solutions of the CBG dominated the solutions of the GPR, resulting in a better selection ability. This comparison highlights the relevance of the additional connections of the CBG, and provides an unique mean to benchmark two different architectures.
\end{itemize}


\section*{Discussion}

The proposed framework listed only the most frequently used data mining algorithms. Other tools can be used depending on the need. Likewise, the examples are provided to illustrate the potential of the approach. Deeper analyses would require to go deeper in the respective fields of these models, what goes beyond the scope of this article. From a methodological point of view, a good practice would be to use this framework to first check in a systematic way a set of well established knowledge from the literature. Further work should then focus on the discrepancies that highlight model errors or incompleteness. A complete agreement to the literature would be a first validation of the model. In a second step, hypotheses currently made can be tested this way by checking their compatibility with the generated data. Although it does not represent a validation as strong as direct observations of the modeled phenomenon, it might help when such observations are not possible or when their cost is so high that a lot of evidence is required to justify such experiments. Although it is somewhat reassuring to follow a systematic path guided by a well defined goal, such analyses is probably also very valuable when little is known in the literature. It allows to propose new hypotheses and then also new experiments to be done on the modeled phenomenon to validate them. 



We have proposed different ways to check whether the set of optimization runs did converge or not.
The goal is to reduce the impact of the stochastic aspect of evolutionary algorithms. A too wide variability indicates that the run did not converge, suggesting that other runs with different algorithm parameters should be performed. Likewise, a finer analysis of the disparity between the runs may reveal area of the objective space for which the optimization was more difficult. Although it is not possible to give a simple and systematic interpretation of this symptom, one of the possibilities is model instabilities and identifying such area of instability may be of interest for the expert performing the analysis. One may wonder about the consequence of the lack of convergence proof on the potential interest of the method. Despite all efforts put into the verifications mentioned before, the convergence to the global optimum will never be sure. Is it a problem for the proposed methodology? Actually, the goal of the multi-objective analysis is to find a set of "interesting" solutions to study. No matter how such points are generated, they are nothing more than specific parameter sets of the model. It means that, even if they are not optimal, they are representative of some model behaviors and are then worth studying. The only risk associated to the non convergence towards the global Pareto-optimal set, is that there might be other behaviors that the analysis will miss. The consequence is that the conclusions of the analysis should rely on the observed behaviors only. Without other hypothesis on model compatibility with optimization, nothing should be deduced about the capacity of the model to generate a non observed behavior, as this absence may stem from a premature convergence towards a local optimum and not from a particular failure of the model. 


Likewise, the proposed comparison between models evaluates at the same time the capacity for the model to optimize the given objectives, but also its compatibility with the optimization algorithm. We might think about a model which is particularly difficult to optimize, for instance if its performance, as defined by the objectives, decreases very fast as its parameters get away from optimal values. Optimizing such a model corresponds to finding a needle in a haystack and in this case, optimization will surely converge towards a local and easier to find optimum. 
It is then possible that a comparison performed as described above will conclude that such a model is less efficient than another one, which is easier to optimize and converges towards an optimum that is above the local optimum of the first model, but maybe not above its global optimum. To be rigorous then, while performing such a comparison, it is necessary to make the hypothesis that both models are compatible with the optimization, i.e. that their global optimum is not on a sharp peak of performance. Although this is difficult to check in general, in practice, such an hypothesis can be tested on well known parameter sets used to study the model. If small changes in these sets results in a performance that is very low (for instance of the same order of magnitude than that of randomly generated sets of parameters), then it is clear that such an hypothesis doesn't hold. In the other case, the hypothesis is unfortunately not validated but can be considered, by default, as reasonable.

Finally, the motivation of this work was to propose an approach to tackle the epistemic opacity inherent to at least some computational models. At this point, we may wonder if we succeeded in making such models less opaque. We have proposed the use of multi-objective optimization algorithms that rely on heuristics, in particular on an inspiration from natural selection and with no strong mathematical grounding. In a second step, we have presented data mining tools to extract from the generated data new knowledge. Doesn't it contribute to make the conclusions weak because of an opaque process? Even if the generation of the data relies on the evolutionary optimization, at the end, what are considered are just some sets of parameters of the model. The method used to generate them is of little impact on the conclusions, as long as no definite conclusions are drawn on the absence of observation of some model behaviors. The evolutionary optimization can then be considered as black box heuristics to discover set of parameters that are worth studying. In this respect, they are not less justified than a manual exploration, as an expert may be driven by biases that are no more justified. Actually, the exploration made by the evolutionary algorithm has the advantage of not being biased by any a priori knowledge on the model. On the other side, an expert, as it can reasonably observe only a small number of model behaviors, will be strongly biased towards parameter values that seem intuitively interesting to him and this intuition may be misleading. The second step, i.e. the analysis step, is not different from current scientific approach were a scientist has to dig in huge amount of data to extract the knowledge he is looking for. In that perspective and from an epistemological point of view, nothing new is proposed here. The opacity of this part of the knowledge building process, if any, is then similar to that of any other experimental science. The epistemic opacity has then not increased. Was it reduced? The analysis highlights potentially unforeseen features of the model, thus leading to a better understanding of its inner working and as a consequence to a reduction of the underlying epistemic opacity, at least for the aspects on which the analysis was focused.

\section*{Conclusion}

We have proposed a framework for studying computational models relying on the use of multi-objective optimization algorithms. The main principle of the methodology consists in exploiting the ability to simulate the behavior of such models a huge number of times to generate model parameter sets that are worth studying. The choice of these data is driven by the optimization of multiple conflicting objectives at the same time, resulting in a set of optimal solutions rather than a single optimal solution. Multi-objective evolutionary algorithms were used for their versatility and for their ability to generate a dense approximation of the set of the best trade-off solutions. Analysis to perform were suggested together with typical questions they may contribute to answer. Two different examples were used to illustrate the potential of the approach: the study of a simulated flapping wing aircraft and the study of basal ganglia brain nuclei models. On the flapping wing aircraft model, the method has highlighted results compatible with the literature for medium to high speeds and non realistic results for low speeds, suggesting to focus further work on this particular point. Likewise, three kinematic parameters revealed to be critical for the minimum energy solutions, opening the way to a reduced model design. On the basal ganglia brain nuclei, the method revealed that the most biologically plausible solutions are also those that are the closest to the functionality attributed to this brain structure. The method also highlighted the importance of a particular connection, whose importance is debated in the neuroscience community and thus suggesting important new hypothesis on its potential use. Finally, a comparison of two brain models using the multi-objective method was performed and the model including the most recent neuroanatomical knowledge dominated the other, thus confirming the importance of the new modeled structures. For the two models, all these results were directly deduced after an analysis of the data generated by the multi-objective evolutionary algorithm.

\section*{Acknowledgements}

This project was partially funded by the ANR EvoNeuro project, ANR-09-EMER-005-01. Hamdaoui M. has been supported by the Systematic cluster through the CSDL project (DGT 117407) (\url{http://www.systematic-paris-region.org/fr/projets/csdl}). 

\small

\bibliographystyle{apalike}

\bibliography{basalganglia,Multi-objective_Analysis,biblio}

\end{document}